\documentclass[10pt,twocolumn,letterpaper]{article}

\usepackage[pagenumbers]{cvpr} 

\usepackage{graphicx}
\usepackage{amsmath}
\usepackage{amssymb}
\usepackage{booktabs}
\usepackage{algorithm}
\usepackage{algorithmic}
\usepackage{tabularx}
\usepackage{enumitem}
\usepackage{lipsum}
\usepackage[pagebackref,breaklinks,colorlinks]{hyperref}

\newcolumntype{C}[1]{>{\centering\arraybackslash}p{#1}}

\newif\ifcomments
\ifcomments
    \newcommand{\david}[1]{\textcolor{red}{[DJ: #1]}}
    
    \newcommand{\daniel}[1]{\textcolor{green}{[DL: #1]}}
    \newcommand{\jiaye}[1]{\textcolor{magenta}{[J: #1]}}
    \newcommand{\soumyadip}[1]{\textcolor{blue}{[SS: #1]}}
\else
    \providecommand{\david}[1]{}
    \providecommand{\daniel}[1]{}
    \providecommand{\jiaye}[1]{}
    \providecommand{\soumyadip}[1]{}
\fi

\usepackage[capitalize]{cleveref}
\crefname{section}{Sec.}{Secs.}
\Crefname{section}{Section}{Sections}
\Crefname{table}{Table}{Tables}
\crefname{table}{Tab.}{Tabs.}

\def\cvprPaperID{5441} 
\def\confName{CVPR}
\def\confYear{2022}

\begin{document}

\title{Fast Light-Weight Near-Field Photometric Stereo}

\author{Daniel Lichy$^1$ \hspace{30pt} Soumyadip Sengupta$^2$ \hspace{30pt} David W. Jacobs$^1$ \\
$^1$University of Maryland, College Park \hspace{30pt} $^2$University of Washington\\
{\tt\small dlichy@umd.edu, soumya91@cs.washington.edu, djacobs@cs.umd.edu}}
\maketitle

\begin{abstract}
\vspace{-0.5em}
We introduce the first end-to-end learning-based solution to near-field Photometric Stereo (PS), where the light sources are close to the object of interest. This setup is especially useful for reconstructing large immobile objects. Our method is fast, producing a mesh from 52 512$\times$384 resolution images in about 1 second on a commodity GPU, thus potentially unlocking several AR/VR applications. Existing approaches rely on optimization coupled with a far-field PS network operating on pixels or small patches. Using optimization makes these approaches slow and memory intensive (requiring 17GB GPU and 27GB of CPU memory) while using only pixels or patches makes them highly susceptible to noise and calibration errors. To address these issues, we develop a recursive multi-resolution scheme to estimate surface normal and depth maps of the whole image at each step. The predicted depth map at each scale is then used to estimate `per-pixel lighting' for the next scale. This design makes our approach almost 45$\times$ faster and 2$^{\circ}$ more accurate  (11.3$^{\circ}$ vs. 13.3$^{\circ}$ Mean Angular Error) than the state-of-the-art near-field PS reconstruction technique, which uses iterative optimization. 

\vspace{-1em}

\end{abstract}


\vspace{-1em}
\section{Introduction}
\vspace{-0.5em}
In this work, we introduce a fast light-weight Photometric Stereo (PS) technique for near-field illumination. Photometric Stereo aims to reconstruct object geometry from a sequence of images captured with a static camera and varying light sources. Existing near-field PS approaches are slow and extremely memory intensive. Being fast and light-weight enables users to capture images and process them on their laptop within a few seconds, allowing multiple retakes if needed. This light-weight reconstruction technique can be extremely useful for several AR/VR applications. While our method is primarily developed for calibrated lighting, in line with existing far-field approaches, we also show how our method can be extended to uncalibrated real-world captures by introducing a calibration network.

Near-field PS is often preferred over far-field or distant lighting-based PS for both practical and theoretical reasons. It is extremely useful for capturing large objects, e.g. furniture or humans, especially in a confined space like a room \cite{Liao2017IndoorSR,Ahmad,Papadhimitri}. This is because far-field PS approaches assume the lighting to be distant, e.g. 10$\times$ the object dimensions is suggested by \cite{Xie2015PhotometricSW,woodham1980},  causing it to be unsuitable for 3D imaging in many indoor spaces. Additionally, low-intensity LED lights on handheld devices (e.g. flashlight on a phone) may not be bright enough to illuminate an object from a large distance \cite{Quau2017SemicalibratedNP}. Theoretically, in the case of uncalibrated lighting, near-field PS has no linear ambiguity in contrast to far-field PS where there is the well-known Generalized Bas-Relief ambiguity \cite{GBR}, as shown in \cite{Papadhimitri}.

We make our method fast and accurate by forgoing traditional optimization in favor of a recursive multi-scale algorithm. Our proposed method consists of two recursive networks one for predicting surface normal and another for depth maps. At each step of the recursion, we increase the input image resolution by a factor of 2. We first analytically estimate the relative lighting direction and attenuation factor for each pixel in the image (termed `per-pixel lighting' for clarity) by upsampling the predicted depth map from the previous step. We then infer the surface normal for this scale given the input image, `per-pixel lighting,' and estimated normal map from the previous scale. Finally, the depth map is predicted conditioned on the estimated normal map and the depth map from the previous scale. The number of steps for this recursion is dictated by the input image resolution making the inference extremely fast, requiring only a few forward passes. We also improve inference speed by using a recursive deep network for estimating depth map from normals instead of solving normal integration by e.g. solving the Poisson equation \cite{horn_and_brooks,normal_integration_survey}, making it more robust to outliers during training. The recursion allows the use of one network for all scales, thus heavily reducing the memory footprint. This approach is also more robust to noise and lighting calibration errors than existing per-pixel based methods  \cite{Logothetis2020ACB,santo2020deep} as the recursion leads to a larger receptive field for the network. 

Our method is built on the shoulders of existing near-field and far-field PS techniques by adapting the ideas that can best improve performance, inference speed, and memory requirements. Our recursive approach is inspired by \cite{Lichy_2021_CVPR}, which uses a single network for predicting normal at each scale conditioned on the image and the estimated normal from the previous scale. It is non-trivial to adapt the recursion idea proposed in \cite{Lichy_2021_CVPR} from far-field distant lighting to near-field because per-pixel lighting directions are not known a priori. Our ablation study shows that a trivial extension of \cite{Lichy_2021_CVPR} to near-field PS that does not refine lighting directions based on depth  performs significantly worse (by 3.5$^{\circ}$) than our proposed approach. The idea of using depth map to predict `per-pixel lighting' is inspired by \cite{Logothetis2020ACB,santo2020deep,Papadhimitri}. However, these approaches operate on pixels or patches using iterative optimization, causing extensive memory usage, slow inference speed and making them highly susceptible to noise and lighting calibration errors.

We first evaluate our method quantitatively on the LUCES dataset \cite{LUCES} with calibrated lighting and show that our method is 2$^{\circ}$ more accurate in surface normal prediction (11.3$^{\circ}$ vs. 13.3$^{\circ}$ Mean Angular Error) than state-of-the-art near-field PS approach L20 \cite{Logothetis2020ACB}, and another prior approach S20 \cite{santo2020deep}. In terms of computational efficiency, our method requires 4GB CPU memory and 12GB GPU memory compared to 27GB CPU and 17GB GPU of L20 \cite{Logothetis2020ACB} for 1024$\times$786 resolution, while S20 \cite{santo2020deep} fails to scale up to this resolution. Our inference speed is 1.3 secs compared to 59.5 secs of L20 \cite{Logothetis2020ACB} and 2435 secs of S20 \cite{santo2020deep} for 52 512$\times$384 resolution images; tested on the same hardware. 

For many practical applications, such as quickly reconstructing 3D models at the home, calibrated lighting is impractical. In the absence of calibrated lighting, we also introduce an additional lighting calibration network. We first show that on the LUCES dataset with uncalibrated lighting our method is more robust than existing approaches, producing 14.11$^{\circ}$ Mean Angular Error (MAE) vs 18.85$^{\circ}$ of L20 and 16.03$^{\circ}$ of S20. Finally, we capture a few real-world objects with near-field lighting with a commodity flashlight and show that our reconstructed mesh is qualitatively more accurate than existing approaches S20~\cite{santo2020deep} and L20~\cite{Logothetis2020ACB}, after using the same calibration network, see Fig. \ref{fig:teaser} and \ref{fig:real_data}.

In summary our contributions are as follows:
\begin{itemize}[noitemsep,topsep=0pt,leftmargin=*]
    \item A state-of-the-art, fast, light-weight, near-field PS method with 45$\times$ faster inference speed and significantly lower memory requirements than existing methods.
    
    \item We build on \cite{Lichy_2021_CVPR}, developed for far-field PS, by incorporating `per-pixel lighting', adding recursive depth prediction from normal, and allowing the flexibility to use unstructured lighting.

    \item We also introduce a calibration network to facilitate uncalibrated capture in-the-wild with an iPhone camera and a handheld flashlight.

\end{itemize}
\vspace{-0.5em}
\section{Prior Work}
\vspace{-0.5em}

Research on Photometric Stereo (PS), introduced in \cite{woodham1980}, can be divided along a number of dimensions: diffuse vs. specular materials, calibrated vs. uncalibrated lighting, distant vs. nearby lights. In this work, we focus on near-field PS with both known and unknown lighting conditions.

\textbf{Far-Field Photometric Stereo.} We briefly mention some recent far-field PS works that are particularly relevant to this work. For a more comprehensive survey see \cite{ackermann2015survey,durou2020advances}. Our work is inspired by \cite{Lichy_2021_CVPR} which introduces a recursive neural net to predict surface normal at each scale given the input image at that scale and the predicted normal map from the previous scale. The authors showed that using a recursive architecture significantly improves performance by capturing global context that is often absent in per-pixel techniques \cite{Ikehata_2018_ECCV} and patch-based techniques \cite{chen2019SDPS_Net}.

\textbf{Near-Field Photometric Stereo.}

Solutions to near-field PS can be roughly divided into two broad approaches. 

The first approach relies on a three step iterative refinement \cite{Bony2013TridimensionalRB,Nie2016ANC,Collins20123DRI,Papadhimitri,Logothetis2020ACB,Ahmad,Quau2017SemicalibratedNP}, starting with an initial shape, e.g. a plane, until convergence:
(1) based on the current shape calculate the light directions and intensity at each point; 
(2) using these light estimates, predict surface normals; 
(3) integrate normals to update the shape. Logothetis \textit{et al.}~\cite{Logothetis2020ACB} uses a per-pixel far-field deep neural network in step (2) while the rest of these methods are purely optimization driven. In contrast, we use two deep recursive neural nets for steps (2) and (3), trained on the whole image for near-field lighting.

Direct optimization approaches rely on inverting the image formation process, often by solving a system of PDEs \cite{Q18,Mecca2014NearFP,Quau2017SemicalibratedNP,Quau2016UnbiasedPS_PDE,Mecca2016ASP_PDE}. 
For a detailed discussion of these methods see \cite{Q18}. In \cite{Xie2015PhotometricSW} the authors use a local-global mesh deformation scheme to optimize a mesh that reconstructs the images. Santo \textit{et al.} \cite{santo2020deep} also optimizes a reconstruction loss. However, as part of their forward pass they decompose observations into reflectance and normal using a far-field deep neural network.

\textbf{Light Calibration.} Research on uncalibrated PS either separately estimates lighting or alternately solves for light and shape simultaneously using a variational approach \cite{variational_light_opt}. For the former, the lighting estimation can be physically performed by inserting additional objects \cite{Liao2017IndoorSR,hertzman_example_obj} in the scene or by using a deep network for prediction \cite{chen2019SDPS_Net,Kaya_2021_CVPR,chen2020learned}. While the these methods have been introduced for far-field PS, we propose a calibration network for near-field PS.

\textbf{Normal Integration.} Normal integration techniques estimate a depth map that is consistent with a normal map. For a detailed discussion see \cite{normal_integration_survey}. Ho \textit{et al.} \cite{ho_fast_marching} uses the similarity between normal integration and shape from shading (SfS) to develop a normal integration technique. Similarly, we also introduce a deep network for faster and stable normal integration during training based on SfS.

\vspace{-0.5em}
\section{Background}
\vspace{-0.5em}
\label{sec:background}
In this section, we describe our image formation model for near-field Photometric Stereo (PS). Given $M$ images of an object ($I^1,...,I^M$) captured under different known anisotropic point light sources from a fixed viewpoint, we estimate the surface normal and the depth map. Additionally, we assume the camera has known intrinsic parameters, and the mean distance to the object is known (WLOG assume mean distance is 1. See Appendix \ref{app:global_scale} for details). This is the same setup as \cite{Logothetis2020ACB,santo2020deep}. In Sec. \ref{sec:cali}, we show how to remove the restriction on known lights and mean distance.

\begin{table}
\begin{tabularx}{\columnwidth}{@{}p{0.15\textwidth}X@{}}
\toprule
  $R$  & number of resolutions \\
  $r_0,...,r_{R-1}$ & sequence of resolutions $r_0=64$, $r_{i+1} = 2r_i$, $r_{R-1}$ input image resolution \\
  $I^j_i$ & $j$th image at resolution $r_i$ \\
  $N_i$ & normal at resolution $r_i$ \\
  $D_i$ & depth at resolution $r_i$ \\
  $A_i^j,L_i^j$ & per-pixel light attenuation and direction at resolution $r_i$ for image $j$\\
  $p^j,d^j,\mu^j$  & light parameters of $j$th image \\
  $Up(I)$ & upsample $I$ by a factor of 2 \\
  $\text{ones}(r\times r)$ & $r\times r$ array of ones \\
\bottomrule
\end{tabularx}
\vspace{-1em}
\caption{Summary of major notations used throughout the text.}
\label{tb:notation}
\vspace{-1.5em}
\end{table}

\textbf{Camera Model} 
We use the standard pinhole camera model centered at the origin in world coordinates and looking down the z-axis. The camera is specified by a 3$\times$3 intrinics matrix $K$. Any world point $X=(x,y,z)$, projects onto a pixel $(u,v)$ by the formula:
\vspace{-0.5em}
\begin{equation}
\label{eq:camera}
(u,v,1)^T \sim K (x,y,z)^T.
\vspace{-0.5em}
\end{equation}

\textbf{Geometry Model} We only consider reconstructing the visible region of an object. Therefore the object is completely described by a normal and depth map. 
Concretely, $X(u,v) \in \mathbb{R}^3$ describes a point on the object appearing in pixel $(u,v)$. Then we can define the depth map by $D(u,v) = X(u,v)_3$, where the subscript 3 refers to the 3rd i.e. $z$ component of $X(u,v)$. We can also recover $X(u,v)$ from the depth map $D(u,v)$ following eqn. \ref{eq:depth_to_param}:

\vspace{-0.5em}
\begin{equation} 
\label{eq:depth_to_param}
X(u,v) = D(u,v)K^{-1}(u,v,1)^T
\vspace{-0.5em}
\end{equation}

If $n(X)$ is the normal at the point $X$ then the normal map is defined by $N(u,v) = n(X(u,v))$. Since $X(u,v)$ is a parametrization, we can also calculate the normal map as:
\vspace{-0.5em}
\begin{equation}
\label{depth_to_normal}
    N = \frac{(\frac{\partial X}{\partial u} \times \frac{\partial X}{\partial v})}{\|(\frac{\partial X}{\partial u} \times \frac{\partial X}{\partial v})\|}.
    \vspace{-0.5em}
\end{equation}

\textbf{Light Model}
We assume each image $I^j$ is illuminated by an anisotropic point light source. We describe this light by a position $p^j \in \mathbb{R}^3$, a direction  $d^j \in S^2$, and an angular attenuation coefficient $\mu^j \in \mathbb{R}$. We assume all lights have unit intensity. If that is not the case, we divide the image by the intensity of the light sources. 

We can then describe the direction of the light arriving at a point $X$ on the surface of the object by:
\vspace{-0.5em}
\begin{equation}
\label{eq:light_direction}
    L^j(X) = \frac{(X-p^j)}{\|X-p^j\|},
    \vspace{-0.5em}
\end{equation}

\noindent and the attenuation of the light at the same point by:
    \vspace{-0.5em}
\begin{equation}
\label{eq:light_attenuation}
   A^j(X) =  \frac{ (L^j \cdot d^j )^{\mu^j} }{ ||X-p^j||^2}.
       \vspace{-0.5em}
\end{equation}

Thus lighting at any pixel $(u,v)$, given the depth map $D(u,v)$, can be described by a direction term $L^j(X(u,v))$ and an intensity attenuation term $A^j(X(u,v))$ (where $X$ is expressed with depth $D$ by eqn. \ref{eq:depth_to_param}). To keep it concise, we term these lighting factors, relative direction and attenuation, at each pixel \textbf{`per-pixel lighting'}.

\textbf{Admissible lights} The configuration of possible anisotropic point lights is huge, taking 3+2+1 parameters to describe. To remedy this we restrict ourselves to lights with positions in a cylinder around the camera and direction pointing roughly toward the object. We term this region the `admissible light region'. It covers positions of lights used in most existing datasets (e.g. \cite{LUCES,santo2020deep}) and  the uncalibrated data we capture. For the exact specification of the admissible light region please see Appendix \ref{app:admissible_region}.

\textbf{Reflectance Model}
We model the reflectance as a general spatially varying BRDF that depends on the lighting direction $\omega_l$, the viewing direction $\omega_v$ and the position on the surface $X$. Denote this as $B(\omega_l,\omega_v,X)$. 

\textbf{Rendering Equation}
Now given the depth map D, normal N, camera intrinsics K, and light parameters $p^j$,$d^j$,$\mu^j$, we can write the rendering equation for the $j$th image as a function of $(u,v)$:
    \vspace{-0.5em}
\begin{equation}
 \label{rendering_equation}
 \small{
 I^j(u,v) =  A^j(X) B(\omega_v, L^j(X))  (N(u,v) \cdot  L^j(X)) + \eta(u,v)}
     \vspace{-0.5em}
 \end{equation} 
 where $\eta$ represents indirect lighting effects such as shadows and inter-reflections. Note that $\omega_v=-X / \|X\|$ because the camera is centered at the origin.

\vspace{-0.5em}
\section{Our Approach}
\vspace{-0.5em}

We aim to predict normal map $N$ and depth map $D$, given a set of images $I^1,...,I^M$. We propose a recursive solution to this problem. We introduce two recursive networks, one for predicting normal $G_{RN}(\cdot;\theta_{RN})$ and another for predicting depth $G_{RD}(\cdot;\theta_{RD})$ given normals. At each step of the recursion we increase the image resolution by a factor of two and use these two networks to predict the depth map and the normal map. For a robust and accurate normal estimation, we calculate the `per-pixel lighting' ($L^{j}$ and $A^{j}$) and use it as an input to the normal estimation network, which we ablate in Sec. \ref{sec:ablation}.

Lichy \textit{et al.}~\cite{Lichy_2021_CVPR} introduced a similar recursive normal estimation network, RecNet, for far-field PS. They showed that the recursive network has a large receptive field and produces high-quality reconstruction by refining the predictions from the previous scale. We also find this idea to be suitable to produce fast and light-weight inference. Thus we developed our own version of recursively reconstructing the object for near-field PS which we describe in Sec. \ref{sec:recursive}. Network architecture, training data and loss functions are described in detail in Sec. \ref{sec:implementation}.

The key differences between our approach and \cite{Lichy_2021_CVPR} are:
\begin{itemize}[noitemsep,topsep=0pt,leftmargin=*]
\item We create synthetic data for training that emulates near-field capture with lighting in the admissible region.
\item We calculate the per-pixel lighting and use it as an extra input to the recursive normal estimation network, which improves performance by 3.5$^{\circ}$, as shown in ablation study (Sec. \ref{sec:ablation}).
\item We introduce a recursive normal to depth integration network, which is fast and robust during training. Predicted depth map is then used for calculating the per-pixel lighting in the next scale.
\item Unlike RecNet, which requires a fixed sequence of lights, our method is permutation invariant to lighting order and can use arbitrary lighting within  the admissable region.
\end{itemize}

\vspace{-0.5em}
\subsection{Recursive Reconstruction}
\vspace{-0.5em}

\label{sec:recursive}

We first initialize the recursion with input resolution of $r_{0}=64\times64$.

\begin{itemize}[noitemsep,topsep=0pt,leftmargin=*]
  \item We first calculate the `per-pixel lighting' parameters $L^{j}_{0}(X)$ and $A^{j}_{0}$ by assuming the depth map is a plane at depth 1 (see, Sec. \ref{sec:background} and Appendix \ref{app:global_scale}). This calculation is done following Algo. \ref{algo:per_pixel_light_dirs_near}.
  
  \item Then we use an initial normal estimation network $G_{IN}(\cdot;\theta_{IN})$, which takes the input image and the per-pixel lighting parameters to predict the normal map $N_{0}$:
  \vspace{-0.5em}
  \begin{equation}
      N_0 = G_{IN}(\{I_{0}^j,L_{0}^j, A_{0}^j\}_{j=1}^M) ; \theta_{IN})
  \vspace{-0.5em}
  \end{equation}
  
  \item Finally, we introduce another initialization network to predict a depth from the normals: $D_0 = G_{ID}(N_0;\theta_{ID})$
\end{itemize}

The recursive network progressively increases input image resolution by a factor of 2, until it reaches the input image resolution. The steps of the recursive network are in principle similar to the initialization network, except for the fact that the normal and depth estimation networks $G_{IN}$ and $G_{ID}$ do not use any recursion and simply predict at low resolution in a feed-forward fashion. The steps of the recursion are explained below:
\begin{itemize}[noitemsep,topsep=0pt,leftmargin=*]
\item For each step $i$ with resolution $r_{i} \times r_{i}$, we first calculate the per-pixel lighting (Algo. \ref{algo:per_pixel_light_dirs_near}) using the depth map of the previous scale $D_{i-1}$ upsampled by a factor of 2.

\item Then normal map $N_i$ is predicted with the recursive normal prediction network $G_{RN}(\cdot;\theta_{RN})$, given the input images and per-pixel lighting along with depth map $D_{i-1}$ and normal map $N_{i-1}$ of the previous scale following:
\vspace{-0.5em}
\begin{equation}
      N_i = G_{RN}(\{I_{i}^j,L_{i}^j, A_{i}^j\}_{j=1}^M, N_{i-1}, D_{i-1} ; \theta_{RN})
  \end{equation}
  
\item Finally we predict the depth map $D_i$ from the normal map $N_i$ using another recursive network $G_{RD}(\cdot;\theta_{RD})$, which is conditioned on the depth map of the previous scale $D_{i-1}$: $D_i = G_{RD}(N_i, D_{i-1};\theta_{RD})$
  
\end{itemize}

The forward pass of our recursive process is also summarized in Algo. \ref{algo:near_field_algorithm}.

\vspace{-0.5em}
\begin{algorithm}[H]
\caption{Calculate the per-pixel lighting given depth D.}
\label{algo:per_pixel_light_dirs_near}
\begin{algorithmic}[1]
\STATE $\textbf{PPLight}(K,D, \mu, p, d)$
\STATE $X[u,v] = D[u,v] K^{-1} (u,v,1)^T$
\STATE $L[u,v] = normalize(X[u,v] - p)$
\STATE $A[u,v] = \frac{(L[u,v] \cdot d)^{\mu}}{||x[u,v] - p||^2}$
\RETURN $A, L$
\end{algorithmic}
\end{algorithm}

\setlength{\textfloatsep}{0pt}
\vspace{-1.5em}
\begin{algorithm}[H]
\caption{Forward pass of our approach: See Tab. \ref{tb:notation} definition of the notation.}
\label{algo:near_field_algorithm}
\begin{algorithmic}[1]

\STATE $L_{0}^j, A_{0}^j = \text{PPLight}(K, \text{ones}(r_0\times r_0), \mu^j, p^j, d^j)$
\STATE $N_{0} = G_{IN}( \{(I_{0}^j,L_{0}^j, A_{0}^j)\}_{j=1}^M; \theta_{IN})$
\STATE $D_{0} = G_{ID}( N_{0}; \theta_{ID})$
\FOR {i = 1 to R-1} 
    \STATE $L_i^j, A_i^j = \text{PPLight}(K,Up(D_{i-1}),\mu^j, p^j, d^j)$
    \STATE $N_{i} = G_{RN}(  \{(I_{i}^j,L_{i}^j, A_{i}^j)\}_{j=1}^M, N_{i-1}; \theta_{RN})$
    \STATE $D_{i} = G_{RD}(N_{i},D_{i-1}; \theta_{RD})$
\ENDFOR
\end{algorithmic}
\end{algorithm}

\vspace{-1em}

\subsection{Implementation Details}
\vspace{-0.5em}

\label{sec:implementation}

\textbf{Network Architectures}. Our method consists of four neural networks, two for initialization and two for recursion with similar architectures for initialization and recursion.

The normal estimation networks consist of a shared encoder that takes in each image $I_i^j$ concatenated with its per-pixel lighting maps $A_i^j$ and $L_i^j$ and returns a feature $F_i^j$ with dimension $128$ at 1/4'th of the input resolution. In the recursion step, the normal from the previous step bilinearly upsampled by a factor of 2 is used as additional input. Then we perform a max pooling operation over the features $F_i^j$s from all input images to produce a combined feature, which is passed to a decoder to produce a normal map.

The depth prediction network takes in the normal estimated by the normal prediction network (in the recursive case the encoder takes in the depth from the previous step bilinearly upsampled by a factor of 2) and produces a depth map. It also does some preprocessing to correct for a perspective camera. Specifically, it applies a transformation (e.g., see \cite{normal_integration_survey} or Appendix \ref{app:perspective_correction}) so that in the perspective case normal integration amounts to solving $\nabla u = (p,q)$ where $u$ is the logarithm of depth and $p$, $q$ are determined by the normal map and camera intrinsics. Architecturally, it is an encoder-decoder ResNet architecture similar to \cite{Lichy_2021_CVPR}. Details can be found in Sec. \ref{sec_normal_integration_net} and Appendix \ref{app:network_architectures}.

\textbf{Loss Function.}
We train our network with three loss functions. We use depth loss $L_{depth}$ and normal loss $L_{normal}$ to produce accurate reconstruction. We also use a loss to ensure the normals derived from the predicted depth map are consistent with those derived from the ground truth depth map. This loss is necessary to produce smooth depth maps. We term this loss $L_{nfd}$, \textit{nfd} is an abbreviation for `normal from depth'.  The losses are defined as:

\vspace{-2em}
\begin{align} 
L_{depth} &= {\textstyle\sum}_{i=0}^{R-1} ||D_i - \bar{D}_i ||_1,\\
L_{normal} &= {\textstyle\sum}_{i=0}^{R-1} ||N_i - \bar{N}_i ||_1,\\
L_{nfd} &= {\textstyle\sum}_{i=0}^{R-1} ||nfd(D_i) - nfd(\bar{D}_i) ||_1,
\end{align}
\noindent where we use a bar above a letter to indicate the Ground Truth (GT) measurement. $nfd$ is the function that takes a depth map and produces a normal map. This is implemented using eqn. \ref{eq:depth_to_param} and \ref{depth_to_normal}. In eqn. \ref{depth_to_normal} we approximate the derivatives with a central finite difference.

\textbf{Training Details.}
Our network is trained completely on synthetic data. First we generate depth, normal, spatially-varying albedo, and Cook-Torrance roughness maps using 14 objects from the statue dataset \cite{sculpture_data} and freely available albedo maps from \cite{3dtextures}. These are rendered at 512$\times$512 resolution. At training time, for each normal, depth, albedo, and roughness, 10 lights are uniformly randomly sampled from the admissible region \ref{sec:background}. With a 50\% probability, we replace the object's material with one from the MERL dataset \cite{Merl}. We then render the 10 images using eqn. \ref{rendering_equation}. 

For augmentation, we randomly zero patches and add random noise to each pixel to simulate the indirect lighting term $\eta$ in eqn. \ref{rendering_equation}. Images are also randomly cropped to simulate a diverse set of camera intrinsics. We trained our network end-to-end for 22 epochs using the Adam optimizer with learning rate 0.0001. Training took about 2 days on 4 Nvidia P6000 GPUs.

\vspace{-0.5em}

\subsection{Normal Integration Network}
\label{sec_normal_integration_net}
\vspace{-0.5em}

We found that existing normal integration algorithms are too slow for use during training of a neural network. Additionally, they fail on our challenging synthetic data due to large discontinuities. Our solution is to replace a classical normal integration routine with a network, but this is a non-trivial task. Solving normal integration requires global information (details in Appendix \ref{app:global_scale}), but convolutional networks have limited receptive fields, and therefore cannot take global information into account for large enough images.

RecNet, a recursive architecture introduced in \cite{Lichy_2021_CVPR}, creates a convolutional network with potentially infinite receptive field. We found a straight forward application of RecNet fails for normal integration. We believe this has to do with the relation between normals and depth. To understand this, we look at the opposite problem i.e. we want to train a network to predict normals from depth. To keep things simple let's consider the orthographic case in 1D, where estimating the normal is the same as estimating the derivative. 

We consider an image as discrete samples of a function on domain $\left[0,1\right]$. Let $0=x_1,...,x_r=1$ be the sample points and let $h = 1/r$ be the distance between them, where $r$ is the image resolution. Let $u$ be the depth and $u_i=u(x_i)$. Let $u'$ be the derivative of $u$ and $  \left[u'\right]_i = u(x_i)$. Let $\{u_i\}$ indicate the sequence of all the elements $u_i$.

Suppose we train a fully convolutional network to predict normal $\{\left[u'\right]_i\}$ from depth $\{u_i\}$ at a resolution $r$. It will learn something similar to a finite difference and return $\{\frac{u_{i+1}-u_{i-1}}{h}\}$. Now if we test the network on an image $\{v_i\}$ that has a higher resolution say  e.g. $2r$. Then the network will predict $\{\frac{v_{i+1}-v_{i-1}}{h}\}$, but this is not the desired result. The correct result is $\{\frac{v_{i+1}-v_{i-1}}{h/2}\}$, this is because the network does not know the resolution has changed. In this case, there is a simple solution:  predict $\{u_i\}$ from the resolution independent $\{\left[u' \right]_i \cdot h \}$ instead of $\{\left[u'\right]_i\}$

This suggests that when we solve the inverse problem we should try to learn a function $G$ that takes $\{\left[u' \right]_i \cdot h\}$ and predicts $\{u_i\}$: $\{u_i\} = G(\{\left[u' \right]_i \cdot h\})$.
This is impossible for a fully convolutional network because it requires global information. However, if we already know a low-resolution estimate of $\{u_i\}$, termed $\{w_j\}$, we then learn a function:
\vspace{-0.5em}
\begin{equation}
   \{ u_i \} = G(\{\left[u' \right]_i \cdot h\}, \{w_j\}),
   \label{eq:normal_int_single_step}
\end{equation}
\noindent i.e. we predict depth from normal and a low resolution estimate of depth. We argue that this is possible for a fully convolutional network. By applying eqn. \ref{eq:normal_int_single_step} recursively, we can gradually reconstruct a full resolution depth map. This is the essential idea of our depth prediction network. For more on this argument and the depth prediction network please see Appendix \ref{app:normal_prediction}.

\vspace{-0.5em}

\subsection{Lighting calibration}
\label{sec:cali}
\vspace{-0.5em}

Setting up calibrated lights in-the-wild is very challenging. Recent works have shown that in the far-field case lighting calibration can be accomplished with a neural network \cite{chen2019SDPS_Net,Kaya_2021_CVPR}. We are not aware of any learning based approach to near-field lighting calibration. 

 Since, in the near-field case, there is much more freedom of possible light configurations, we make some additional simplifying assumptions on the light:  (1) the light intensity is the same in all images.  (2) the light is well modeled by an isotropic point source i.e $\mu=0$ and $d$ is irrelevant. (3) The light is within the admissible region.  We found that these assumptions are good enough to estimate light from a handheld flashlight used for capture in-the-wild.

We use essentially the same architecture as  \cite{chen2019SDPS_Net} to estimate light positions. This network uses a shared feature extractor to extract a feature $F^j$ from each image $I^j$. It then creates a context $c= \text{max}_j F^j$. Finally, a second network is applied to feature $F_j$ and the context $c$ to produce a light position estimate $p^j$ for image $I^j$. To deal with a perspective camera, all input images are cropped or zero padded to have the same intrinsics.

\vspace{-0.5em}

\section{Experimental Evaluation}
\vspace{-0.5em}

We evaluate our method quantitatively on the LUCES dataset \cite{LUCES} in Sec. \ref{sec:luces}, and qualitatively on a dataset we captured with a handheld flashlight and iPhone in Sec.~\ref{sec:real}. We mainly compare our results to two state-of-the-art near-field Photometric Stereo (PS) algorithms S20 \cite{santo2020deep} and L20 \cite{Logothetis2020ACB}. In the case of uncalibrated capture, we use our calibration network described in Sec. \ref{sec:cali} for S20 and L20, which are only developed for calibrated lighting conditions.

\begin{figure}
    \centering
    \includegraphics[width=0.95\columnwidth]{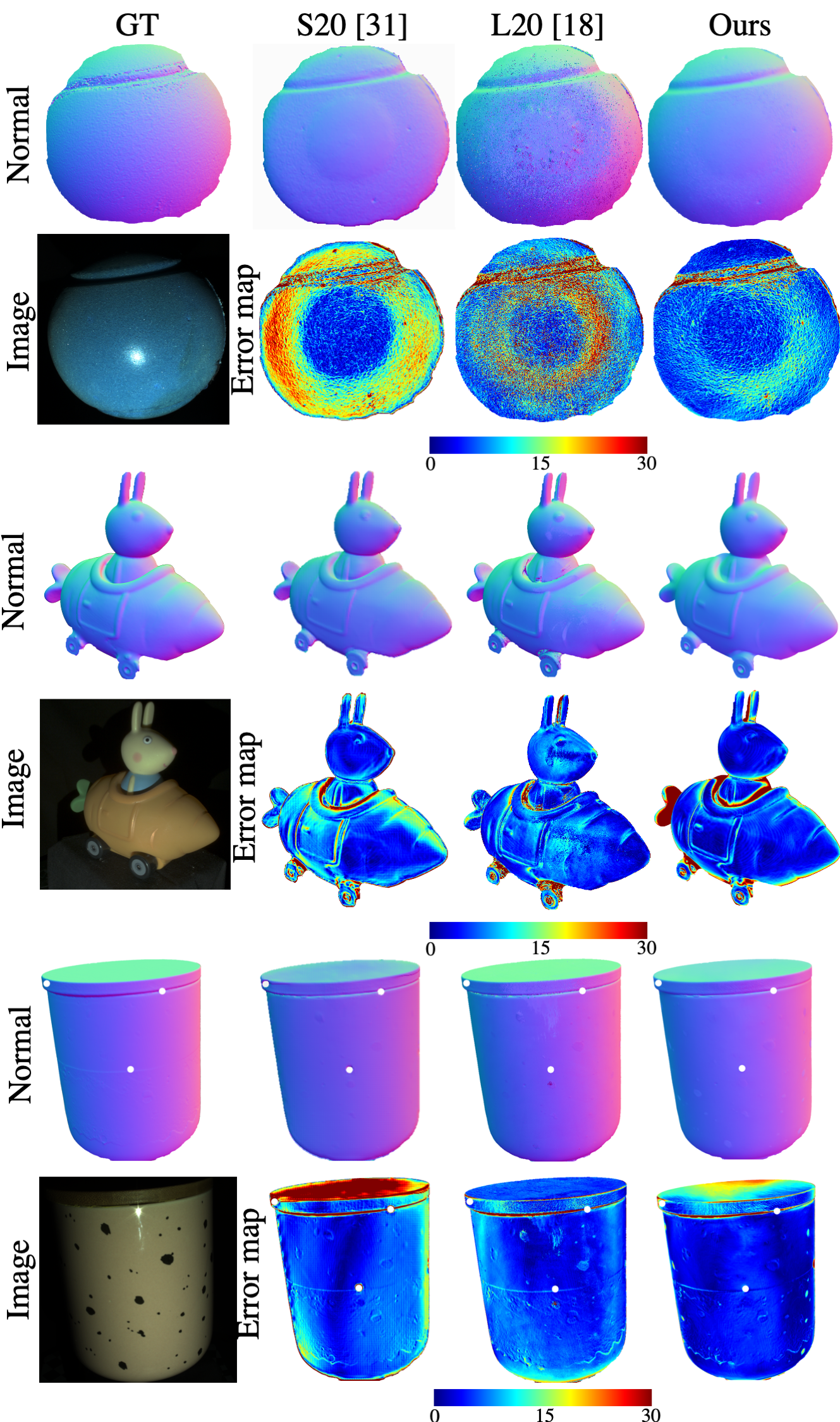}
    \vspace{-0.5em}
    \caption{We compare the predicted normal map and an error map w.r.t. GT of our approach with that of S20~\cite{santo2020deep} and L20~\cite{Logothetis2020ACB} on sample objects from the LUCES \cite{LUCES} with calibrated lighting.}
    \label{fig:luces_cal}
\end{figure}

\begin{table*}[t]
\begin{center}
\resizebox{2.0\columnwidth}{!}{%
\hspace{-0.0125\columnwidth}

\begin{tabular}{ | c | c c c c c c c c c c c c c c c c|}
 \hline
Method & Error & Bell & Ball & Buddha & Bunny & Die & Hippo & House & Cup & Owl & Jar & Queen & Squirrel & Bowl & Tool & Average \\ \hline
 L17-\cite{Logothetis_2017_CVPR} & MAE & 28.25 & 9.77 & 11.5 & 20.15 & 11.95 & 15.42 & 29.69 & 30.76 & 13.77 & 10.56 & 13.05 & 15.93 & 12.5 & 15.1 & 17.03 \\ 
 & MZE & 4.45 & 0.81 & 4.67 & 7.51 & 4.58 & 3.19 & 6.99 & 2.67 & 3.64 & 6.56 & \textbf{1.89} & 1.82 & 4.37 & 3.25 & 4.02 \\  \hline
 Q18-\cite{Q18} & MAE & 25.8 & 12.12 & 14.07 & 13.73 & 13.77 & 18.51 & 30.63 & 37.63 & 14.74 & 15.66 & 13.16 & 14.06 & 11.19 & 16.12 & 17.94 \\ 
 & MZE & 12.03 & 2.5 & 9.28 & 7.06 & 5.91 & 6.8 & 8.02 & 4.83 & 5.83 & 16.87 & 6.92 & 2.55 & 6.48 & 6.69 & 7.27 \\   \hline
S20-\cite{santo2020deep} & MAE & 9.5 & 25.42 & 19.17 & 12.5 & 5.23 & 23.12 & \textbf{28.02} & \textbf{14.22} & 13.08 & 9.27 & 16.62 & 14.07 & 12.44 & 17.42 & 15.72 \\
 & MZE & 1.9 & 5.5 & 5.53 & 6.02 & 2.76 & 7.04 & \textbf{6.15} & \textbf{1.62} & 3.75 & 6.09 & 3.91 & 2.81 & 5.22 & 4.68 & 4.5 \\ \hline
L20-\cite{Logothetis2020ACB} & MAE & 14.74 & 12.43 & \textbf{10.73} & \textbf{8.15} & 6.55 & \textbf{7.75} & 30.03 & 23.35 & 12.39 & 8.6 & \textbf{10.96} & 15.12 & 8.78 & 17.05 & 13.33 \\
 & MZE &  \textbf{1.53}	& 	\textbf{0.67}	& 	\textbf{3.27}	& 	2.49	& 	4.44	& 	\textbf{1.82}	& 	9.14	& 	2.04	& 	3.44	& 	\textbf{3.86}	& 	1.94	& \textbf{1.01}	& 	\textbf{2.80}	& 	5.90	& 	3.17 \\ \hline 
I18-\cite{Ikehata_2018_ECCV} & MAE & 23.55 & 44.29 & 35.29 & 36 & 41.52 & 44.9 & 49.05 & 35.78 & 40.27 & 40.66 & 32.89 & 41.09 & 28.04 & 31.71 & 37.5 \\
 & MZE & 5.93 & 6.59 & 10.92 & 6.88 & 7.83 & 7.59 & 8.98 & 3.17 & 8.67 & 15.54 & 8.08 & 5.8 & 6.69 & 12.45 & 8.22 \\ \hline
Ours & MAE & \textbf{6.20} & \textbf{8.55} & 12.69  & 8.63  & \textbf{5.16} & 8.01 & 29.00 & 17.28 & \textbf{12.32} & \textbf{5.32} & 12.90 & \textbf{13.00}   & \textbf{7.07} & \bf{12.33} & \textbf{11.32} \\
 & MZE & 2.28 & 1.83 & 16.60  & 2.73  & 2.76 & 3.52 & 7.39  & 2.00  & 3.08  & 6.58 & 3.09 & 3.94     & 3.78 & 2.48 & 4.43 \\
 & MZE int & 1.71 & 1.26 & 3.93   & \textbf{2.16}  & \textbf{2.16} & 3.44 & 7.07  & 1.82  & \textbf{2.85}  & 3.99 & 2.90 & 1.88     & 3.61 & \textbf{2.27} & \textbf{2.93} \\ \hline \hline
GT & Diff-MAE & 2.5 & 2.69 & 2.69 & 2.93 & 2.49 & 3.2 & 9.19 & 2.85 & 4.3 & 1.79 & 4.22 & 3.26 & 2.27 & 2.34 & 3.34 \\
& Int\cite{Quau2015EdgePreservingIO}-MZE & 0.08 & 	0.22 & 	3.28 & 	2.30&  0.56& 1.28& 	7.43 & 	0.02	& 	3.51	&  0.12	& 	3.25 & 	1.12 & 	0.12 & 	0.13 & 	1.67 \\ \hline
\end{tabular}
} 
\end{center}
\vspace{-1.5em}
\caption{\small{Evaluation on LUCES dataset with calibrated lighting. Mean angular error (MAE in degrees) and mean depth error (MZE in mm).}} 
\label{table:luces_cal}
\vspace{-1.0em}
\end{table*}

\begin{figure}[!h]
    \centering
    \includegraphics[width=0.95\columnwidth]{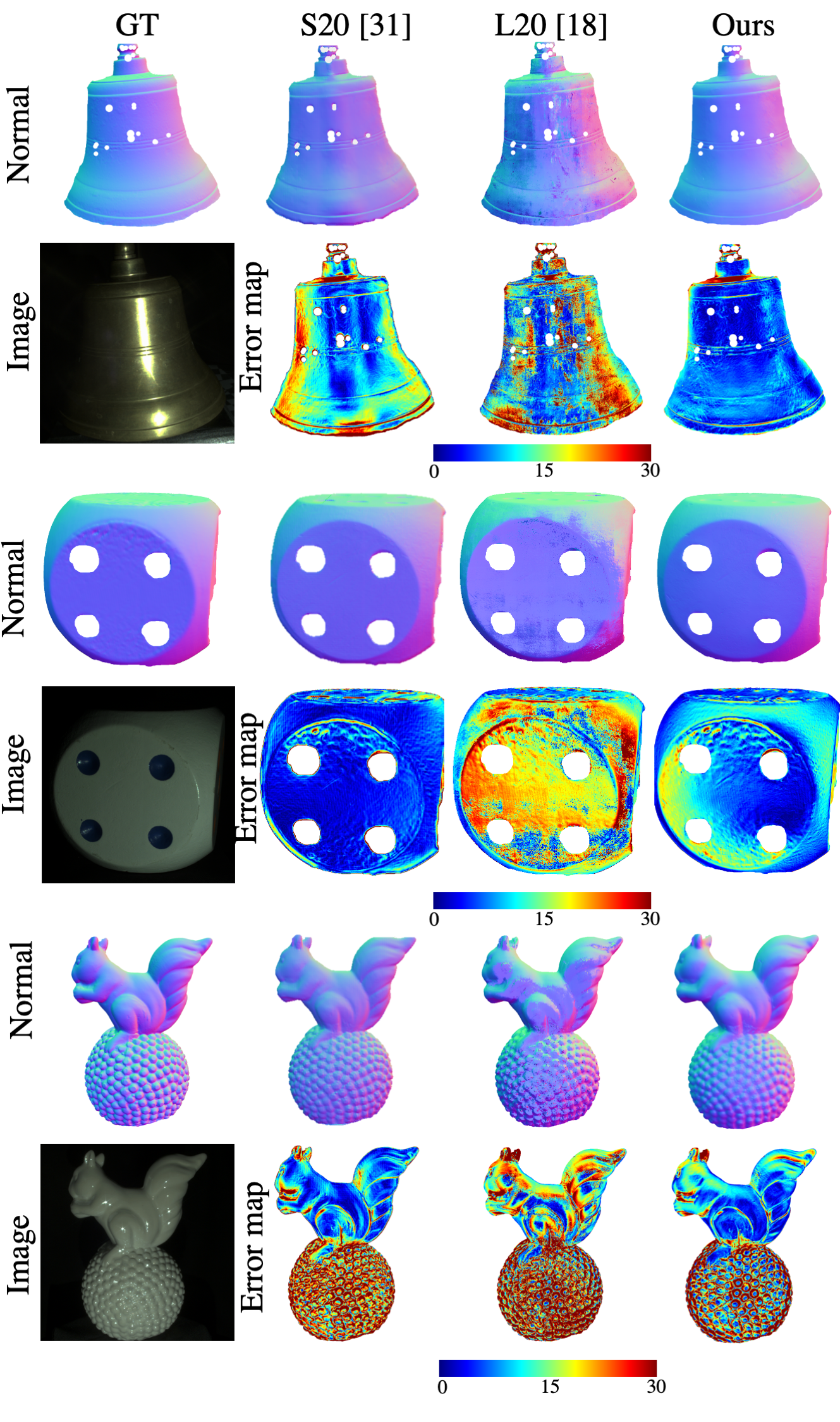}
    \vspace{-1em}
    \caption{We compare the predicted normal map and an error map w.r.t. GT of our approach with that of S20~\cite{santo2020deep} and L20~\cite{Logothetis2020ACB} on sample objects from the LUCES \cite{LUCES} with uncalibrated lighting.}
    \label{fig:luces_uncal}
\end{figure}

\begin{table*}[!ht]
\begin{center}
\resizebox{2.0\columnwidth}{!}{%
\hspace{-0.0125\columnwidth}

\begin{tabular}{ | c | c c c c c c c c c c c c c c c c|}
 \hline
Method & Error & Bell & Ball & Buddha & Bunny & Die & Hippo & House & Cup & Owl & Jar & Queen & Squirrel & Bowl & Tool & Average \\ \hline

 S20-\cite{santo2020deep} & MAE & 13.43  & 13.68 &	21.85 &	\textbf{11.41}	& \textbf{5.86}	& 11.33	& 36.28	& \textbf{17.63}	& 17.67	& 11.60 &	\textbf{15.96} &	17.24		& 13.92	& \textbf{16.54}	& 16.03 \\
 & MZE & 2.98	& 2.58	& 8.48	& 4.91	& 2.52	& \textbf{2.85}	& 10.31 &	\textbf{1.95} &	6.44 &	6.18 &	4.85 &	3.18 &		4.39 &	\textbf{2.38} &	4.57\\ \hline
 L20-\cite{Logothetis2020ACB} & MAE & 13.97 &	15.50 &	18.92 &	14.71 &	16.14 &	16.20 &	32.06 &	23.80	& 17.81	 & 17.65 &	20.79	& 21.45 &		11.83 &	23.00	& 18.85 \\
  & MZE & 4.21 &	3.21 &	10.39	& 4.88	& 5.58	& 4.29	&  12.04	&  2.18	& 5.59	& 10.14 &	8.27 &	3.26	&	3.32 &	6.12	& 5.96 \\ \hline
 Ours & MAE & \textbf{7.17} &	\textbf{6.59}	& \textbf{14.50} &	11.75	& 8.63	& \textbf{10.64}	& \textbf{31.00} &	18.98	& \textbf{15.92}	& \textbf{9.14}	& 18.39	& \textbf{15.97} &		\textbf{10.17} &	18.61	& \textbf{14.11} \\
 & MZE & \textbf{1.80} &	1.40	& 10.27	& 3.84 &	2.77	& 3.59	& 10.04	& 2.64	& \textbf{4.13}	& 7.35 &	\textbf{4.59} &	3.51 &		\textbf{3.15} & 	6.93	& 4.71 \\
 & MZE int & 2.32 &	\textbf{0.76}	& \textbf{5.52} &	\textbf{2.79} &	\textbf{2.41} &	3.32 &	\textbf{8.68} &	2.07 &	4.77 &	\textbf{4.97} &	5.15 &	\textbf{2.60} &		3.59 &	6.08 &	\textbf{3.93} \\ \hline

\end{tabular}
} 
\end{center}
\vspace{-1.5em}
\caption{\small{Evaluation on LUCES with uncalibrated lighting. Mean angular error (MAE in degrees) and mean depth error (MZE in mm).}}
\label{table:luces_uncal}
\vspace{-1.5em}
\end{table*}

\vspace{-0.5em}

\subsection{Quantitative Evaluation on LUCES \cite{LUCES}}
\label{sec:luces}
\vspace{-0.5em}

The LUCES dataset consists of 14 objects, each captured in HDR under 52 calibrated near-field lighting conditions. We evaluate using the mean angular error (MAE) for normal and mean depth error (MZE) metrics.

\textbf{Calibrated.} In Tab. \ref{table:luces_cal}, we present MAE and MZE obtained by our method and compare it with existing works as reported in LUCES \cite{LUCES}. The table includes results from two pure optimization near-field methods L17 \cite{Logothetis_2017_CVPR} and Q17 \cite{Q18}, two hybrid near-field methods using deep learning and optimization S20 \cite{santo2020deep} and L20 \cite{Logothetis2020ACB}, and the far-field deep method I18 \cite{Ikehata_2018_ECCV}. All methods are evaluated at the 2048$\times$1536 resolution, except S20 which was evaluated at 512$\times$384 due to its GPU memory requirements \cite{LUCES}.

Tab. \ref{table:luces_cal} shows our method outperforms all existing methods in terms of MAE, especially the state-of-the-art method L20 (MAE 13.33$^\circ$ vs. 11.32$^\circ$). Using our integration network we are the second best in MZE, L20 outperforming us by 1.26mm. We found that our normal integration network can develop jumps at discontinuities (see Appendix \ref{app:limitations} for details), which increase our MZE.  We tried  resolving this issue by integrating our normal map predictions as a post processing step. We used the optimization approach of \cite{Quau2015EdgePreservingIO}, also used by L20 for this step. After this post processing, our MZE dropped to 2.93mm, and our method becomes the top performer. We label our method with this post processing step as MZE+int in Tab. \ref{table:luces_cal}. It is tempting to try replacing our normal integration network with a traditional integration algorithm during training, but we find these algorithms struggle with our synthetic training data due to discontinuities in the data. Furthermore, these methods are very slow for use during network training. 

Tab. \ref{table:luces_cal} also reports the MAE achieved by differentiating the ground truth depth map with a finite difference (Diff-MAE) and the depth error obtained by integrating the ground truth normal with \cite{Quau2015EdgePreservingIO} (Int-MZE). These errors are due to the discrete nature of images and discontinuities in the object, see \cite{LUCES} for detailed discussion.

We show normal prediction results and error maps from each of these methods in Fig. \ref{fig:luces_cal}. Depth error visualizations are included in Appendix \ref{app:additional_results}.

\textbf{Uncalibrated.} In Tab. \ref{table:luces_uncal} we compare our method to L20 and S20 where the ground truth lighting calibration is replaced with the results of our calibration network. Since our calibration network only handles the case of equal intensity lights we scale each image by their ground truth intensity. Additionally lights are assumed to be isotropic point sources i.e. we set $\mu^j=0$ in each method. Our method and L20 were evaluated at 1024$\times$768 resolution. S20 was evaluated at 512$\times$384 resolution due to GPU memory limitations. Results of S20 were then bilinearly upsampled to 1024$\times$768 for error evaluation.

We again observe that our method is the best in MAE (14.11$^\circ$ vs. 16.03$^\circ$ for S20). Using our normal integration network we are slightly surpassed by S20 in MZE (4.57mm vs. 4.71mm). However, after using post-processing normal integration we improve our MZE to 3.93mm. Results and error maps are shown in Fig. \ref{fig:luces_uncal}.

\vspace{-0.5em}

\subsection{Computational Resources}
\vspace{-1em}

\label{sec:comp-eval}

\begin{table}[!h]
\begin{center}
\begin{tabular}{ |c|c|c|c|c|}
\hline

method & res. & time(s) & cpu (GB) & gpu (GB)\\ \hline
S20-\cite{santo2020deep} & 512 &	2435.0	&	5	&	20\\ 
L20-\cite{Logothetis2020ACB} & 512 &	59.5	&	8	&	5\\ 
Ours & 512 &	1.3 (2.0) &	4	&	9\\ \hline 
L20-\cite{Logothetis2020ACB} & 1024 &	200.0	&	27	&	17\\ 
Ours & 1024 &	4.0 (6.9) &	4	&	12\\ \hline 
\end{tabular}
\end{center}
\vspace{-1.5em}
\caption{Comparison of computational resources. Our method produces significantly faster inference while consuming less CPU and GPU memory than S20 and L20. The quantities in brackets for our method indicate post-processing normal integration. S20 cannot operate on 1024 resolution (res) due to memory limitations.}
\label{table:comp_resources}
\vspace{-1em}
 \end{table}

We compare the memory usage and inference speed of our method to that of S20 and L20 in Tab. \ref{table:comp_resources}. All methods were tested on the same machine with a 24GB Nvidia P6000 GPU and 128GB of main memory. We compare our method and L20 at two resolutions 512$\times$384 and 1024$\times$768. S20 is only compared at 512$\times$384 due to its GPU memory requirements.

\textbf{Inference Speed} We calculate inference speed without the time required for data reading and writing, which can vary depending on the cluster load. Our method is 45$\times$ faster than the closest competitor L20 at both $512$ and $1024$ resolution. Adding the normal integration post-processing step to our method increases the runtime by about 50\%, still leaving our method 30$\times$ faster than L20. S20 is over 1000$\times$ slower than our method.

\textbf{CPU memory} The amount of CPU memory used by S20 and our method are essentially fixed for a given number of input images. In contrast, L20 uses CPU memory approximately proportional to image resolution. 

\textbf{GPU memory} At 512 resolution S20 requires 20GB of GPU memory to process LUCES, see Tab. \ref{table:comp_resources}. This was the highest resolution we were able to run on a 24GB GPU and is consistent with that reported in \cite{LUCES}. GPU usage for L20 and our method are more moderate, however there are some subtleties that must be taken into account to fairly compare GPU usage. See Appendix \ref{app:gpu_memory} for details.

\begin{table}[!h]
    \centering
    \vspace{-1em}
    \begin{tabular}{|c|c|c|}
    \hline
       & MAE & MZE \\ \hline
     with per-pixel lighting estimation & \textbf{11.32} & \textbf{4.43} \\
     w/o per-pixel lighting estimation & 14.88 & 4.89 \\
     \hline
    \end{tabular}
    \vspace{-0.5em}
    \caption{We show that using per-pixel lighting as input to the recursive normal prediction network improves reconstruction accuracy.}
    \label{tab:ablation}
    \vspace{-0.5em}
\end{table}

\vspace{-1.5em}

\subsection{Qualitative Comparison on Captured Data}
\label{sec:real}
\vspace{-0.5em}

We captured a dataset of medium to large size objects using an iPhone12 mounted on a tripod and a handheld flashlight. We capture a short video (5-10 second) of each object while moving the flashlight around the admissible region. Every fifth frame from the video was then used as input images. Images were inverse tonemapped by raising them to the power 2.2. Lighting positions were estimated with our calibration network. 

We compare our method with that of S20~\cite{santo2020deep} and L20~\cite{Logothetis2020ACB}, presented in Fig. \ref{fig:real_data} and \ref{fig:teaser}, with more in Appendix \ref{app:additional_results}. L20's performance is strongly affected by noise due to its per-pixel normal prediction network. S20's predicted normal maps show strong checkerboard patterns.

\begin{figure}
    \centering
    \includegraphics[width=\columnwidth]{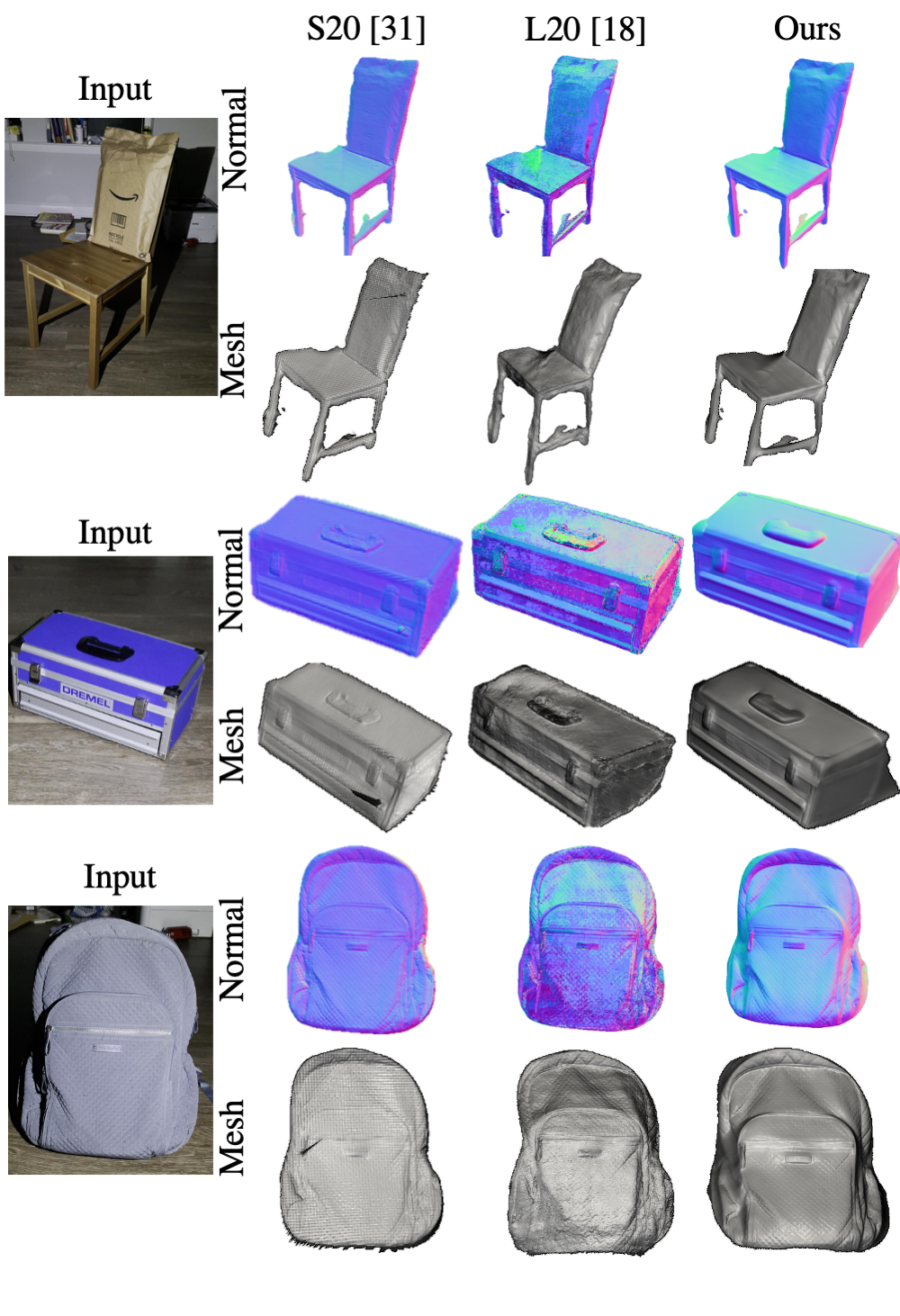}
    \vspace{-3.5em}
    \caption{Qualitative evaluation on images captured with a handheld flashlight and iPhone 12 camera mounted on a tripod. Our method outperforms L20 and S20.}
    \label{fig:real_data}
    \vspace{-0em}
\end{figure}

\vspace{-1em}

\subsection{Ablation Study}
\label{sec:ablation}
\vspace{-0.5em}

\textbf{Importance of Per-Pixel Lighting} We train a variant of our network that does not require per-pixel lighting estimation as input to the normal estimation network. Concretely, rather than using the per-pixel lighting $A^j_i$ and $L^j_i$ as input to the network $G_{IN}$ and $G_{RN}$ we only input the lights' parameters $p^j,d^j,\mu^j$ at each pixel. Note that, these light parameters are associated with global lighting conditions and do not reflect per-pixel lighting effects unlike $A^j_i$ and $L^j_i$. We observe that explicitly estimating and using per-pixel lighting as input improves the performance from 14.88$^\circ$ to 11.32$^\circ$ MAE. This also shows that a more direct adaptation of the recursion idea presented by Lichy \textit{et al.} in \cite{Lichy_2021_CVPR} from far-field to near-field PS, is less effective without using per-pixel lighting estimation. We discuss this result in more depth in Appendix \ref{app:ablation}.

\vspace{-0.5em}
\section{Conclusion}
\vspace{-0.5em}
In this work, we introduce a fast light-weight solution to near-field Photometric Stereo. Existing approaches rely on optimization with point or patch based inferences, which are more susceptible to noise, lack global context, and are memory intensive and slow.  The key innovation of this work lies in creating a system that can capture global context to produce accurate predictions while being light-weight and fast. We adapt  \cite{Lichy_2021_CVPR} from far-field PS to near-field by introducing per-pixel lighting estimation, a recursive normal integration network, and extend it to handle arbitrary lighting. We show in an ablation study that a straightforward adaption of \cite{Lichy_2021_CVPR} produces worse results than our approach. We also introduce a calibration network and show how our approach can be used for capturing 3D geometry of large and midsize real-world objects, like furniture, backpacks, etc. Our method significantly outperforms state-of-the-art methods on both the LUCES dataset and real-world captures while being fast and light-weight.

\textbf{Acknowledgment} This research is supported by the NSF under grant no. IIS-1910132.

\newpage
{\small
\bibliographystyle{ieee_fullname}
\bibliography{egbib}

\begin{thebibliography}{10}\itemsep=-1pt

\bibitem{3dtextures}
3d textures.
\newblock \url{https://3dtextures.me/}.
\newblock Accessed: 2020.

\bibitem{ackermann2015survey}
Jens Ackermann and Michael Goesele.
\newblock A survey of photometric stereo techniques.
\newblock {\em Foundations and Trends{\textregistered} in Computer Graphics and
  Vision}, 9(3-4):149--254, 2015.

\bibitem{Ahmad}
Jahanzeb Ahmad, Jiuai Sun, Lyndon Smith, and Melvyn Smith.
\newblock An improved photometric stereo through distance estimation and light
  vector optimization from diffused maxima region.
\newblock {\em Pattern Recognition Letters}, 50:15--22, 2014.

\bibitem{GBR}
P.N. Belhumeur, D.J. Kriegman, and A.L. Yuille.
\newblock The bas-relief ambiguity.
\newblock In {\em Proceedings of IEEE Computer Society Conference on Computer
  Vision and Pattern Recognition}, pages 1060--1066, 1997.

\bibitem{Bony2013TridimensionalRB}
Alexandre Bony, Benjamin Bringier, and Majdi Khoudeir.
\newblock Tridimensional reconstruction by photometric stereo with near spot
  light sources.
\newblock {\em 21st European Signal Processing Conference (EUSIPCO 2013)},
  pages 1--5, 2013.

\bibitem{chen2019SDPS_Net}
Guanying Chen, Kai Han, Boxin Shi, Yasuyuki Matsushita, and Kwan-Yee~K. Wong.
\newblock Sdps-net: Self-calibrating deep photometric stereo networks.
\newblock In {\em CVPR}, 2019.

\bibitem{chen2020learned}
Guanying Chen, Michael Waechter, Boxin Shi, Kwan-Yee~K Wong, and Yasuyuki
  Matsushita.
\newblock What is learned in deep uncalibrated photometric stereo?
\newblock In {\em European Conference on Computer Vision}, 2020.

\bibitem{Collins20123DRI}
Toby Collins and Adrien Bartoli.
\newblock 3d reconstruction in laparoscopy with close-range photometric stereo.
\newblock {\em Medical image computing and computer-assisted intervention :
  MICCAI ... International Conference on Medical Image Computing and
  Computer-Assisted Intervention}, 15 Pt 2:634--42, 2012.

\bibitem{durou2020advances}
Jean-Denis Durou, Maurizio Falcone, Yvain Qu{\'e}au, and Silvia Tozza.
\newblock Advances in photometric 3d-reconstruction, 2020.

\bibitem{variational_light_opt}
Bjoern Haefner, Zhenzhang Ye, Maolin Gao, Tao Wu, Yvain Quéau, and Daniel
  Cremers.
\newblock Variational uncalibrated photometric stereo under general lighting.
\newblock 11 2019.

\bibitem{hertzman_example_obj}
Aaron Hertzmann and Steven~M. Seitz.
\newblock Example-based photometric stereo: Shape reconstruction with general,
  varying brdfs.
\newblock {\em IEEE Trans. Pattern Anal. Mach. Intell.}, 27(8):1254–1264, aug
  2005.

\bibitem{ho_fast_marching}
Jeffrey Ho, Jongwoo Lim, {Ming Hsuan} Yang, and David Kriegman.
\newblock Integrating surface normal vectors using fast marching method.
\newblock In {\em Computer Vision - ECCV 2006, 9th European Conference on
  Computer Vision, Proceedings}, Lecture Notes in Computer Science (including
  subseries Lecture Notes in Artificial Intelligence and Lecture Notes in
  Bioinformatics), pages 239--250, 2006.
\newblock 9th European Conference on Computer Vision, ECCV 2006 ; Conference
  date: 07-05-2006 Through 13-05-2006.

\bibitem{horn_and_brooks}
Berthold~K.P Horn and Michael~J Brooks.
\newblock The variational approach to shape from shading.
\newblock {\em Computer Vision, Graphics, and Image Processing},
  33(2):174--208, 1986.

\bibitem{Ikehata_2018_ECCV}
Satoshi Ikehata.
\newblock Cnn-ps: Cnn-based photometric stereo for general non-convex surfaces.
\newblock In {\em Proceedings of the European Conference on Computer Vision
  (ECCV)}, September 2018.

\bibitem{Kaya_2021_CVPR}
Berk Kaya, Suryansh Kumar, Carlos Oliveira, Vittorio Ferrari, and Luc Van~Gool.
\newblock Uncalibrated neural inverse rendering for photometric stereo of
  general surfaces.
\newblock In {\em Proceedings of the IEEE/CVF Conference on Computer Vision and
  Pattern Recognition (CVPR)}, pages 3804--3814, June 2021.

\bibitem{Liao2017IndoorSR}
Jingtang Liao, Bert Buchholz, Jean-Marc Thiery, Pablo Bauszat, and Elmar
  Eisemann.
\newblock Indoor scene reconstruction using near-light photometric stereo.
\newblock {\em IEEE Transactions on Image Processing}, 26:1089--1101, 2017.

\bibitem{Lichy_2021_CVPR}
Daniel Lichy, Jiaye Wu, Soumyadip Sengupta, and David~W. Jacobs.
\newblock Shape and material capture at home.
\newblock In {\em Proceedings of the IEEE/CVF Conference on Computer Vision and
  Pattern Recognition (CVPR)}, pages 6123--6133, June 2021.

\bibitem{Logothetis2020ACB}
Fotios Logothetis, Ignas Budvytis, Roberto Mecca, and Roberto Cipolla.
\newblock A cnn based approach for the near-field photometric stereo problem.
\newblock {\em ArXiv}, abs/2009.05792, 2020.

\bibitem{Logothetis_2017_CVPR}
Fotios Logothetis, Roberto Mecca, and Roberto Cipolla.
\newblock Semi-calibrated near field photometric stereo.
\newblock In {\em Proceedings of the IEEE Conference on Computer Vision and
  Pattern Recognition (CVPR)}, July 2017.

\bibitem{Merl}
Wojciech Matusik, Hanspeter Pfister, Matt Brand, and Leonard McMillan.
\newblock A data-driven reflectance model.
\newblock {\em ACM Transactions on Graphics}, 22(3):759--769, July 2003.

\bibitem{LUCES}
Roberto Mecca, Fotios Logothetis, Ignas Budvytis, and Roberto Cipolla.
\newblock Luces: A dataset for near-field point light source photometric
  stereo.
\newblock {\em ArXiv}, abs/2104.13135, 2021.

\bibitem{Mecca2016ASP_PDE}
Roberto Mecca, Yvain Qu{\'e}au, Fotios Logothetis, and Roberto Cipolla.
\newblock A single-lobe photometric stereo approach for heterogeneous material.
\newblock {\em SIAM J. Imaging Sci.}, 9:1858--1888, 2016.

\bibitem{Mecca2014NearFP}
Roberto Mecca, Aaron Wetzler, A. Bruckstein, and R. Kimmel.
\newblock Near field photometric stereo with point light sources.
\newblock {\em SIAM J. Imaging Sci.}, 7:2732--2770, 2014.

\bibitem{Nie2016ANC}
Ying Nie, Zhan Song, Ming Ji, and Lei Zhu.
\newblock A novel calibration method for the photometric stereo system with
  non-isotropic led lamps.
\newblock {\em 2016 IEEE International Conference on Real-time Computing and
  Robotics (RCAR)}, pages 289--294, 2016.

\bibitem{Papadhimitri}
Thoma Papadhimitri and Paolo Favaro.
\newblock Uncalibrated near-light photometric stereo.
\newblock {\em BMVC 2014 - Proceedings of the British Machine Vision Conference
  2014}, 01 2014.

\bibitem{Quau2015EdgePreservingIO}
Yvain Qu{\'e}au and Jean-Denis Durou.
\newblock Edge-preserving integration of a normal field: Weighted
  least-squares, tv and l$^1$ approaches.
\newblock In {\em SSVM}, 2015.

\bibitem{Quau2016UnbiasedPS_PDE}
Yvain Qu{\'e}au, Roberto Mecca, and Jean-Denis Durou.
\newblock Unbiased photometric stereo for colored surfaces: A variational
  approach.
\newblock {\em 2016 IEEE Conference on Computer Vision and Pattern Recognition
  (CVPR)}, pages 4359--4368, 2016.

\bibitem{Quau2017SemicalibratedNP}
Yvain Qu{\'e}au, Tao Wu, and Daniel Cremers.
\newblock Semi-calibrated near-light photometric stereo.
\newblock In {\em SSVM}, 2017.

\bibitem{Q18}
Yvain Quéau, Bastien Durix, Tao Wu, Daniel Cremers, Francois Lauze, and
  Jean-Denis Durou.
\newblock Led-based photometric stereo: Modeling, calibration and numerical
  solution.
\newblock {\em Journal of Mathematical Imaging and Vision}, 60, 03 2018.

\bibitem{normal_integration_survey}
Yvain Quéau, Jean-Denis Durou, and Jean-François Aujol.
\newblock Normal integration: A survey.
\newblock {\em Journal of Mathematical Imaging and Vision}, 60, 05 2018.

\bibitem{santo2020deep}
Hiroaki Santo, Michael Waechter, and Yasuyuki Matsushita.
\newblock Deep near-light photometric stereo for spatially varying
  reflectances.
\newblock In {\em European Conference on Computer Vision (ECCV)}, 2020.

\bibitem{sculpture_data}
Olivia Wiles and Andrew Zisserman.
\newblock Silnet : Single- and multi-view reconstruction by learning from
  silhouettes.
\newblock In {\em British Machine Vision Conference 2017, {BMVC} 2017, London,
  UK, September 4-7, 2017}. {BMVA} Press, 2017.

\bibitem{woodham1980}
Robert~J. Woodham.
\newblock {Photometric Method For Determining Surface Orientation From Multiple
  Images}.
\newblock {\em Optical Engineering}, 19(1):139 -- 144, 1980.

\bibitem{Xie2015PhotometricSW}
Wuyuan Xie, Chengkai Dai, and Charlie C.~L. Wang.
\newblock Photometric stereo with near point lighting: A solution by mesh
  deformation.
\newblock {\em 2015 IEEE Conference on Computer Vision and Pattern Recognition
  (CVPR)}, pages 4585--4593, 2015.

\end{thebibliography}
}

\clearpage
\section{Appendix}
\subsection{Overview}
\label{app:overview}

This supplement includes additional details that could not be put into the main paper due to space restrictions. In Sec. \ref{app:global_scale} we define a relative coordinate system (where mean depth is 1) and show how to relate absolute coordinates to this coordinate system. In Sec. \ref{app:admissible_region}, we define the admissible light region in terms of this relative coordinate system. In Sec. \ref{app:normal_prediction}, we discuss the normal integration problem and elaborate more on our depth prediction network. In Sec. \ref{app:network_architectures} we give the low-level details of our network architectures. In Sec. \ref{app:gpu_memory}, we add some additional details regarding GPU memory usage. In Sec. \ref{app:ablation}, we explain why the network without per-pixel lighting from our ablation study performs poorly. In Sec. \ref{app:limitations}, we demonstrate how certain errors can form in our depth prediction network. Finally, in Sec. \ref{app:additional_results}, we present some additional experimental results.

\subsection{Coordinate System and Scale}
\label{app:global_scale}

This section shows how to pick a coordinate system with a mean depth of one and how this resolves the global scale ambiguity in the uncalibrated case. Equations from the main paper Sec. 3 are included below for quick reference.

\vspace{-0.5em}
\begin{equation} 
\label{eq:depth_to_param_app}
X(u,v) = D(u,v)K^{-1}(u,v,1)^T
\vspace{-0.5em}
\end{equation}

\begin{equation}
\label{eq:light_direction_app}
    L^j(X) = \frac{(X-p^j)}{\|X-p^j\|},
    \vspace{-0.5em}
\end{equation}

\begin{equation}
\label{eq:light_attenuation_app}
   A^j(X) =  \frac{ (L^j \cdot d^j )^{\mu^j} }{ ||X-p^j||^2}.
       \vspace{-0.5em}
\end{equation}

\begin{equation}
\label{rendering_equation_app}
 \small{
 I^j(u,v) =  A^j(X) B(\omega_v, L^j(X))  (N(u,v) \cdot  L^j(X)) + \eta(u,v)}
     \vspace{-0.5em}
 \end{equation}

\textbf{Light scale} We assume that all images have the same light intensity. However, we train our model such that the exact value of this intensity is unimportant, i.e. if we multiply all the images by a constant factor, we get the same results. 

We get this effect by dividing each image by the mean intensity of the first image $I^0$ in the input set i.e. $\mu_{intensity} = \text{mean}_{u,v} I^0(u,v)$ then the input to the network is the image set $I^j(u,v)/\mu_{intensity}$.

\textbf{Mean depth} We assume mean depth is known. We can assume the mean depth is one by changing units. In particular, if $\mu_{depth}$ is the mean depth then we can replace $D(u,v)$ with $D^j(u,v)/\mu_{depth}$ and $p^j$ with $p^j/\mu_{depth}$. From equations \ref{eq:depth_to_param_app}, \ref{eq:light_direction_app}, \ref{eq:light_attenuation_app}, and \ref{rendering_equation_app} we see that this just scales the image intensity by $\mu_{depth}^2$, but, as stated above, our network is invariant to the image intensity scale factor.

\textbf{Uncalibrated Scale Ambiguity}
There is a global scale factor ambiguity between the light positions and depth, which is exactly why we can assume the mean depth is one. We train our calibration network on data with mean depth one, so the network predicts lights in this relative coordinate system. This resolves the scale ambiguity.
\subsection{Admissible Light Region}
\label{app:admissible_region}

Now that we have defined our relative coordinate system with mean depth one \ref{app:global_scale}, we can define the admissible light region in terms of it.

We define the admissible light region as a cylinder with its axis along the camera optical axis (a.k.a. z-axis) and radius 0.75. The extent of the cylinder is from 0.15 behind the camera plane to 0.15 in front of the camera plane. Furthermore, we specify the admissible light directions as the directions making an angle of 30$^\circ$ or less with the z-axis.

Note that in absolute units, the size of the admissible region depends on the distance the object is from the camera. For example, if we are capturing a big object, we would place the camera farther away, and thus in absolute units, the admissible region will be larger.
\subsection{Normal Integration}
\label{app:normal_prediction}

In this section, we present the mathematical intuition that inspired our depth prediction network. We then explain the details of the network's application. This section is not particularly rigorous, but we found that the network it inspired works well in practice.  

\subsubsection{High Level Idea}

\textbf{Problem Statement} Give functions $p$ and $q$ on some domain, we want to find a function on the domain satisfying the PDE

\begin{equation}
\label{normal_integration_equation}
\nabla U = (p,q)
\end{equation}

This is equivalent estimating depth from a normal map, see \ref{app:perspective_correction}. In general, a solution $U$ may not exist. In which case, we want to find some approximate solution.

\textbf{Necessity of Global Information} 
We can see that solving eq. \ref{normal_integration_equation} requires global information as follows. Observe that given any solution to eq. \ref{normal_integration_equation} we can obtain another solution by adding a constant to it. Now suppose we broke the domain of interest into patches and produced a solution for each patch. Because each patch solution could have a different offset, we would have to look outside the patch to find the proper offset needed to glue the patch solutions together continuously. Therefore, a standard feed-forward network, which can only look at patches in very high-resolution images due to its limited receptive field, can not generalize to high-resolution data. 

\textbf{Proposed Solution} Suppose we divide the depth into patches as before, but we are given the mean depth of each patch. Then we could solve the equation on each patch and set the patch mean to the given mean, thus producing a solution.

Concretely, consider a rectangular domain. Divide the region into smaller rectangles call them $P_1$,...,$P_N$. Let $\mu_k = \text{mean}_{x \in P_k} U(x)$. Suppose we knew the function $V(x)$ given by
\begin{equation}
    V(x) = \mu_k  \text{ if } x \in P_k 
\end{equation}
i.e. $V$ is constant on the patches.

Now we solve \ref{normal_integration_equation} individually on each patch $P_k$, call the solution $U_k$. Furthermore, choose the $U_k$ such that they have mean value zero. Define 

\begin{equation}
    W(x) =  U_k(x)  \text{ if } x \in P_k
\end{equation}

Then we can produce a solution to \ref{normal_integration_equation} as

\begin{equation}
    U(x) = W(x) + V(x) \text{ if } x \in P_k
\end{equation}

In other words 
\begin{equation}
    \nabla W = \nabla (U(x)-V(x)) = (p,q)-\nabla V(x)
    \label{eq:local_normal_int}
\end{equation}

can be solved by just looking at the individual patches $P_k$. Then $U(x)$ satisfying \ref{normal_integration_equation} can be recovered as $W(x)+V(x)$. Thinking of $V$ as an upsampled version of a low-resolution approximation to $U$ is the motivation for our depth prediction network explained next.

\subsubsection{Depth Prediction Network}
Now we give the details of the depth prediction networks forward pass. The main paper does not distinguish between the preprocessing the depth prediction network does and the convolutional network proper. Here we use $G_{ID}$ and $G_{RD}$ for the networks and preprocessing combined ($G_{*D}$ to refer to both) and $NET_{*D}$ to refer to the convolution nets proper.

Algo. \ref{algo:depth_pred_forward}  gives the forward pass for $G_{*D}$ (in the initial network, $G_{ID}$, the input depth is just a plane at $z=1$). Where $\mathcal{D}$ is the central finite-difference

\begin{equation}
     \mathcal{D}[U] = (U_{(m+1)n} - U_{(m-1)n}, U_{m(n+1)} - U_{m(n-1)})
\end{equation}

Note that in Algo. \ref{algo:depth_pred_forward} line (**) is just the discrete analog of Eq. \ref{eq:local_normal_int}, where we multiply $(p,q)$ by the step size $h_i$ for the reasons explained in the main paper.

\begin{algorithm}[H]
\caption{Forward pass of depth prediction network}
\label{algo:depth_pred_forward}
\begin{algorithmic}[1]
\STATE $G_{*D}(N_i, D_{i-1})$
	\STATE $p_i,q_i =$ from $N_i$ using perspective correction
	\STATE $U_{i-1} = ln(Upsample[D_{i-1}])$
	\STATE $res = NET_{*D}(h_i (p_i,q_i) - \mathcal{D}[ U_{i-1}])$ (**)
	\STATE $U_i = U_{i-1} + res$
	\STATE $D_i = exp(U_i)$
	\RETURN $D_i$
\end{algorithmic}
\end{algorithm}

\subsubsection{Perspective correction}
\label{app:perspective_correction}
Suppose an image is taken with focal length f, depth, $D(u,v)$ and normals, $N(u,v)$. Define $U=ln(D)$ and $p=-\frac{N_1}{u N_1 + v N_2 + f N_3}$ and $q = -\frac{N_2}{u N_1 + v N_2 + f N_3}$. Then $U$,$p$, and $q$ satisfy \ref{normal_integration_equation}.
Therefore, given normals we can make this transformation and solve for $U$ then recover $D$ as $D=exp(U)$. For a derivations see \cite{normal_integration_survey}.

\subsubsection{Alternative Fast Normal Integration Methods}

We considered FFT and DCT based integration methods that are fast enough for network training. However, because these assume a periodic or rectangular domain, they perform poorly on the datasets we tested, which all have irregular domains. For more discussion on this issue please see \cite{normal_integration_survey} Sec. 3.3 and 3.4.

\subsection{Network Architectures}
\label{app:network_architectures}

In this section, we define the low-level architectures of all the networks used in the paper. In Tab. \ref{tb:network_notation} we define the notation 
used to describe network layers.

\begin{table}
\begin{tabularx}{\columnwidth}{@{}p{0.15\textwidth}X@{}}
\toprule
  A-B  & apply layer A then layer B \\
  BN & BatchNorm \\
  Relu(x) & Leaky Relu activation with parameter x if `(x)' is omitted x=0 (i.e. standard Relu)\\
  conv\_kn\_fm\_sp & convolution layer with kernel of size n with m filters and stride p. If stride is 1 we will omit \_s1. \\
  Conv\_kn\_fm\_sp & conv\_kn\_fm\_sp - BN - Relu \\
  ConvL\_kn\_fm\_sp & conv\_kn\_fm\_sp - BN - Relu(0.1) \\
  convt\_kn\_fm\_sp & transposed convolution layer with kernel of size n and m filters  and stride p. \\
  Res\_n & conv\_k3\_fn - BN - Relu - conv\_k3\_fn - BN - +input. This defines the residual block.  +input indicates adding the input value to the output value. \\
  Upsample & bicubic upsampling by a factor of 2 \\
  tanh & hyperbolic tangent activation \\
\bottomrule
\end{tabularx}
\vspace{-1em}
\caption{Definition of notations for network architecture}
\label{tb:network_notation}
\end{table}

\subsubsection{Normal Prediction Network}
Both the initial, $G_{IN}$, and recursive, $G_{RN}$, normal prediction networks have the same architecture. They consist of a shared feature extractor defined by

\begin{itemize}
\item FE = Conv\_k7\_f32 - Res\_32 - Res\_32 - Conv\_k3\_f64\_s2 - Res\_64 - Res\_64 - Conv\_k3\_f128\_s2 - Res\_128 - conv\_k3\_f128
\end{itemize}

and a normal regressor defined by

\begin{itemize}
\item NR = Res\_128 - convt\_k3\_f64\_s2 - BN - Relu - Res\_64 - Res\_64 - convt\_k3\_f32\_s2 - BN - Relu - conv\_k7\_f3
\end{itemize}

The application of the network is given by
\begin{equation}
    N_i = \text{NR}(\max_{j=1}^M \text{FE}(I^j_i,Upsample(N_{i-1}))) + Upsample(N_{i-1})
\end{equation}

Due to the max-pooling the network is invariant to the image ordering.

\subsubsection{Depth Prediction Network}
The architectures of the initial depth prediction network, $G_{ID}$, and the recursive depth prediction network, $G_{RD}$, are the same. They are given by

 \begin{itemize}
\item Conv\_k7\_f32  - Res\_32 - Res\_32 - Conv\_k3\_f64\_s2 - Res\_64 - Res\_64 - Conv\_k3\_f128\_s2 - Res\_128 - Res\_128 - Upsample - Conv\_k3\_f64 - Res\_64 - Res\_64 - Upsample - Conv\_k3\_f32  - conv\_k7\_f1 - tanh
\end{itemize}

The full application of the normal prediction network is defined in Sec. \ref{app:normal_prediction}.
 
 \subsubsection{Light Prediction Network}
 
 The general architecture of the light prediction network is taken from \cite{chen2019SDPS_Net}. Like the normal prediction network, it consists of a feature extractor LFE and a regressor LR defined by
 
\begin{itemize}
\item LFE = ConvL\_k3\_f64\_s2 -  ConvL\_k3\_f128\_s2 - ConvL\_k3\_f128 - ConvL\_k3\_f128\_s2 - ConvL\_k3\_f128 -  ConvL\_k3\_f256\_s2 - ConvL\_k3\_f256
 
\item LR = ConvL\_k3\_f256 - ConvL\_k3\_f256\_s2 - ConvL\_k3\_f256\_s2 - ConvL\_k3\_f256\_s2 
\end{itemize}

It also has three final coordinate regressors CR\_i for each coordinate $x=x_1$, $y=x_2$, $z=x_3$ defined by:

\begin{itemize}
\item CR\_i  = ConvL\_k1\_f64 - convL\_k1\_f1
\end{itemize}

First the network extracts a feature $F^j$ from each image with the LFE network $F^j = \text{LFE}(I^j)$. It then forms a context $c = \max_j F^j$. Then for each feature $F^j$ it applies the LR network to $F^j$ concatenated with the context $c$ to form position features $PF^j = \text{LR}(F^j,c)$. Finally, it applies the coordinate regessor CR\_i to each position feature $PF^j$ to get the light coordinates i.e. the $k$ coordinate of the light position in image j is: 

\begin{itemize}
\item $p^j_k = \text{CR\_i}(PF^j)$
\end{itemize}
\subsection{GPU Memory Usage Details}
\label{app:gpu_memory}

\begin{table}[!h]
\begin{center}
\begin{tabular}{ |c|c|c|c|c|}
\hline

method & res. & time(s) & cpu (GB) & gpu (GB)\\ \hline
S20-\cite{santo2020deep} & 512 &	2435.0	&	5	&	20\\
L20-\cite{Logothetis2020ACB} & 512 &	59.5	&	8	&	5\\ 
Ours & 512 &	1.3 (2.0) &	4	&	9\\ \hline 
L20-\cite{Logothetis2020ACB} & 1024 &	200.0	&	27	&	17\\ 
Ours & 1024 &	4.0 (6.9) &	4	&	12\\ \hline 
\end{tabular}
\end{center}
\vspace{-1.5em}
\caption{Comparison of computational resources. Our method produces significantly faster inference while consuming less CPU and GPU memory than S20 and L20. The quantities in brackets for our method indicate post-processing normal integration. S20 cannot operate on 1024 resolution (res) due to memory limitations.}
\label{table:comp_resources_app}

 \end{table}

This section includes some additional information regarding the GPU memory usage of method L20 and our method.

\textbf{L20} The memory usage of L20 is dependent on the batch size (number of pixels) that the network processes at one time. For our experiments, we used a batch size of 512. GPU memory usage could potentially be reduced by decreasing the batch size.

\textbf{Ours} Despite our method requiring 12GB of GPU memory to run LUCES at 1024 on our cluster, with a Nvidia P6000, we were able to run LUCES at 2048 resolution on a desktop computer with an 8GB Nvidia RTX2080 GPU. This indicates that our method is even more light-weight than indicated in table \ref{table:comp_resources_app}, which is also in the main paper.

\subsection{Ablation Details}
\label{app:ablation}

This section explains why the network without per-pixel lighting performs worse than with per-pixel lighting.

A convolutional neural network essentially applies the same function to each input patch. Suppose we have two patches, one on the left side of the image and one on the right side, that appear the same. In the ablated network with only global light positions as input, the patches look identical, and the network must predict the same normal. However, the lighting direction and intensity at these two patches are really different, so the patches need to be interpreted differently. The network using per-pixel lighting can distinguish between these two patches because they have different per-pixel lighting, and therefore it can produce different accurate results.

\subsection{Limitations}
\label{app:limitations}

\begin{figure}
    \centering
    \includegraphics[width=\columnwidth]{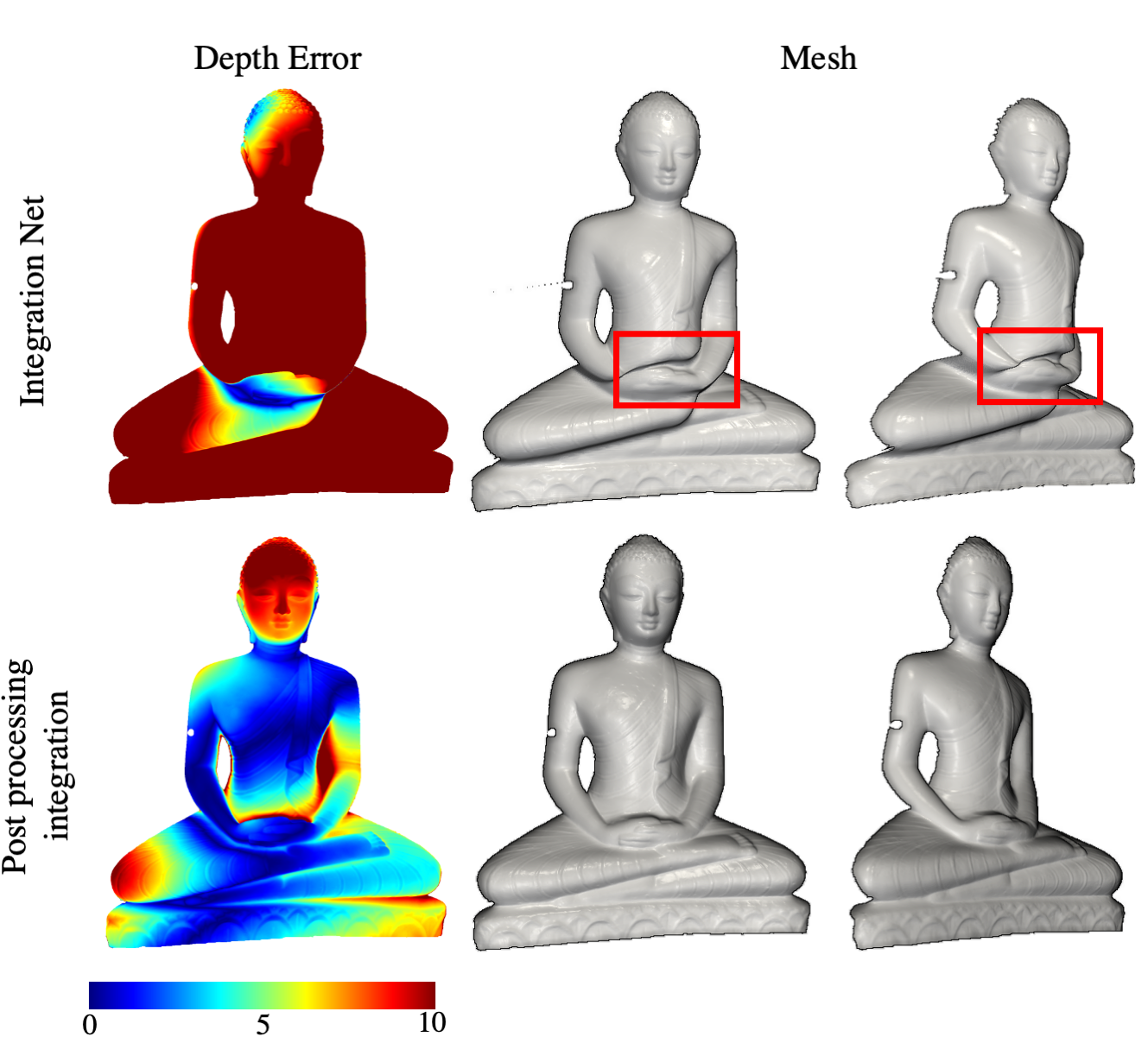}
    \caption{Depth prediction network developing jumps at discontinuities. The jump is highlighted in the red box. Traditional normal integration method \cite{Quau2015EdgePreservingIO} solves this issue (bottom row).}
    \label{fig:jump_discontinuty}
\end{figure}

\textbf{Jumps at discontinuities} As we mention in the main text, our depth prediction method can develop jumps at discontinuities. We show the most extreme example of this type of jump in Fig.  \ref{fig:jump_discontinuty}.

\subsection{Additional Results}
\label{app:additional_results}
This section we presents some additional results and figures.

\textbf{8 Image Results} We compared S20 \cite{santo2020deep}, L20 \cite{Logothetis2020ACB}, and our method on a subset of 8 randomly selected images from the LUCES dataset. They are images: 5, 7, 13, 19, 26, 39, 48, 50. We show the results in  Tab. \ref{table:luces_8_images}. S20's \cite{santo2020deep} code has a bug that caused it to fail with NaN values on two objects: Bell and House. Therefore, we report mean errors for all objects (average 14) for L20 and our methods, and the average over the 12 objects that S20 worked on (average 12).

We observe that our method only drops 1.2$^\circ$ MAE (from 11.32$^\circ$ to 12.44$^\circ$) when tested on this subset. Whereas the other methods drop nearly 3$^\circ$ MAE. 

\begin{table*}[!h]
\begin{center}
\resizebox{2.0\columnwidth}{!}{%
\hspace{-0.0125\columnwidth}

\begin{tabular}{ | c | c c c c c c c c c c c c c c c c c|}
 \hline
Method & Error & Bell & Ball & Buddha & Bunny & Die & Hippo & House & Cup & Owl & Jar & Queen & Squirrel & Bowl & Tool & Average 14 & Average 12\\ \hline

 S20-\cite{santo2020deep} & MAE 	& & 18.08 &	27.2 &	11.69 &	10.06 &	12.58 &	&	23.06 &	13.36 &	14.19 &	16.93 &	18.59 &		12.46	& 15.32 &	& 16.13\\
 & MZE 	& & 2.91 &	6.09 &	3.85 &	3.31 &	\textbf{2.37} &		& 3.62 &	4.52 &	8.83 &	3.30 &	3.13 &		\textbf{3.71} &	3.62 & &	4.10\\ \hline
 L20-\cite{Logothetis2020ACB} & MAE &20.03 &	24.43 &\textbf{	12.67} &	11.85 &	7.18 &	14.12 &	30.74 &	25.63 &	15.72 &	9.22 &	\textbf{13.12} & 15.68	 &		17.88 &	19.01 &	 16.95 & 15.54
 \\
  & MZE & 3.42	& 6.44	& \textbf{4.15} &	 3.20 &	\textbf{1.78} &	3.22 &	8.49 &	3.33 &	5.73 &	4.47 &	4.26 &	2.05 &		8.08 &	9.64 &	4.87 & 4.69 \\ \hline
 Ours & MAE & \textbf{8.84}	& \textbf{9.64} &	13.59	& \textbf{9.31} &	\textbf{5.99} &	\textbf{8.75} &	\textbf{29.43} &	\textbf{21.62} &	\textbf{11.43} &	\textbf{7.13} &	13.38 &	\textbf{13.10}	 &	\textbf{8.73} &	\textbf{13.18}	& \textbf{12.44} & \textbf{11.32}\\
 & MZE & 2.04	& 2.12 &	13.27 &	3.21 &	2.91 &	3.20 &	7.13 &	2.85 &	\textbf{3.51} &	7.84 &	3.06 &	3.68 &		3.84 &	3.04 &	4.41 & 4.38 \\
 & MZE int & \textbf{1.68} &	\textbf{1.46} &	4.70 &	\textbf{2.22} &	2.43 &	3.13 &	\textbf{6.21} &	\textbf{2.23} &	4.01 &	\textbf{4.41} &	\textbf{3.01}	& \textbf{1.97} &		4.09 &	\textbf{3.07} &	\textbf{3.19} & \textbf{3.06} \\ \hline

\end{tabular}
} 
\end{center}

\caption{\small{Evaluation on LUCES with only 8 input images per object with calibrated lighting. Mean angular error (MAE in degrees) and mean depth error (MZE in mm). S20 failed with NaN errors on Bell and House. Average 14 is the average error with all 14 LUCES objects and Average 12 is the average error of the objects excluding Bell and House.}}
\label{table:luces_8_images}
\vspace{-1.5em}
\end{table*}

\begin{figure}[!h]
    \centering
    \includegraphics[width=\columnwidth]{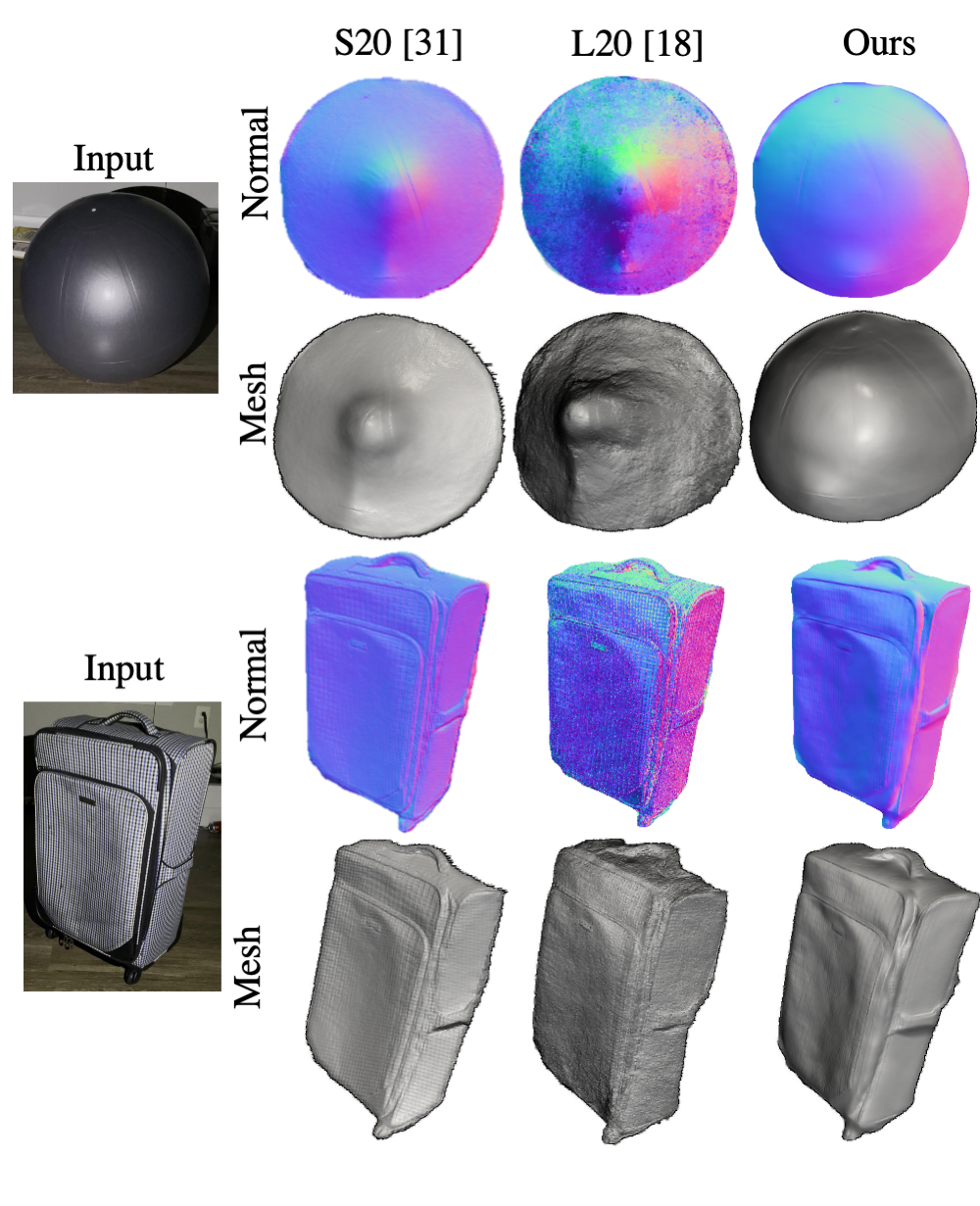}
    \vspace{-8mm}
    \caption{Additional results on data captured by us.}
    \label{fig:itw_capture}
\end{figure}

\textbf{Additional Results on Our Data}
In Fig. \ref{fig:itw_capture}, we compare the results of S20 \cite{santo2020deep}, L20 \cite{Logothetis2020ACB}, and our method on data we captured.

\begin{figure}[!h]
    \centering
    \includegraphics[width=\columnwidth]{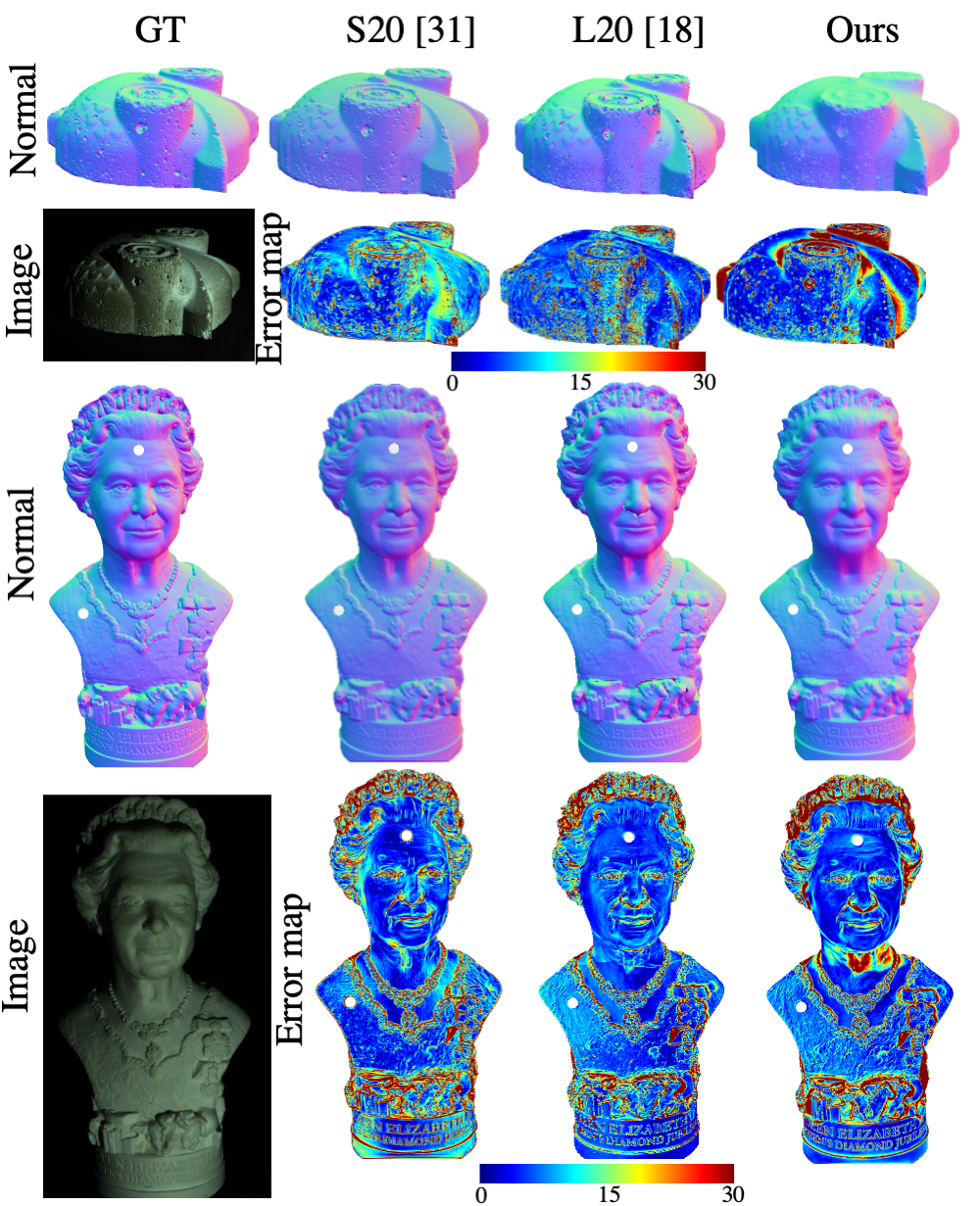}
    \caption{Additional normal predictions and error maps (in degrees) on the LUCES dataset in the calibrated case.}
    \label{fig:normal_err_cal}
\end{figure}

\begin{figure*}[!h]
    \centering
    \includegraphics[width=0.9\textwidth]{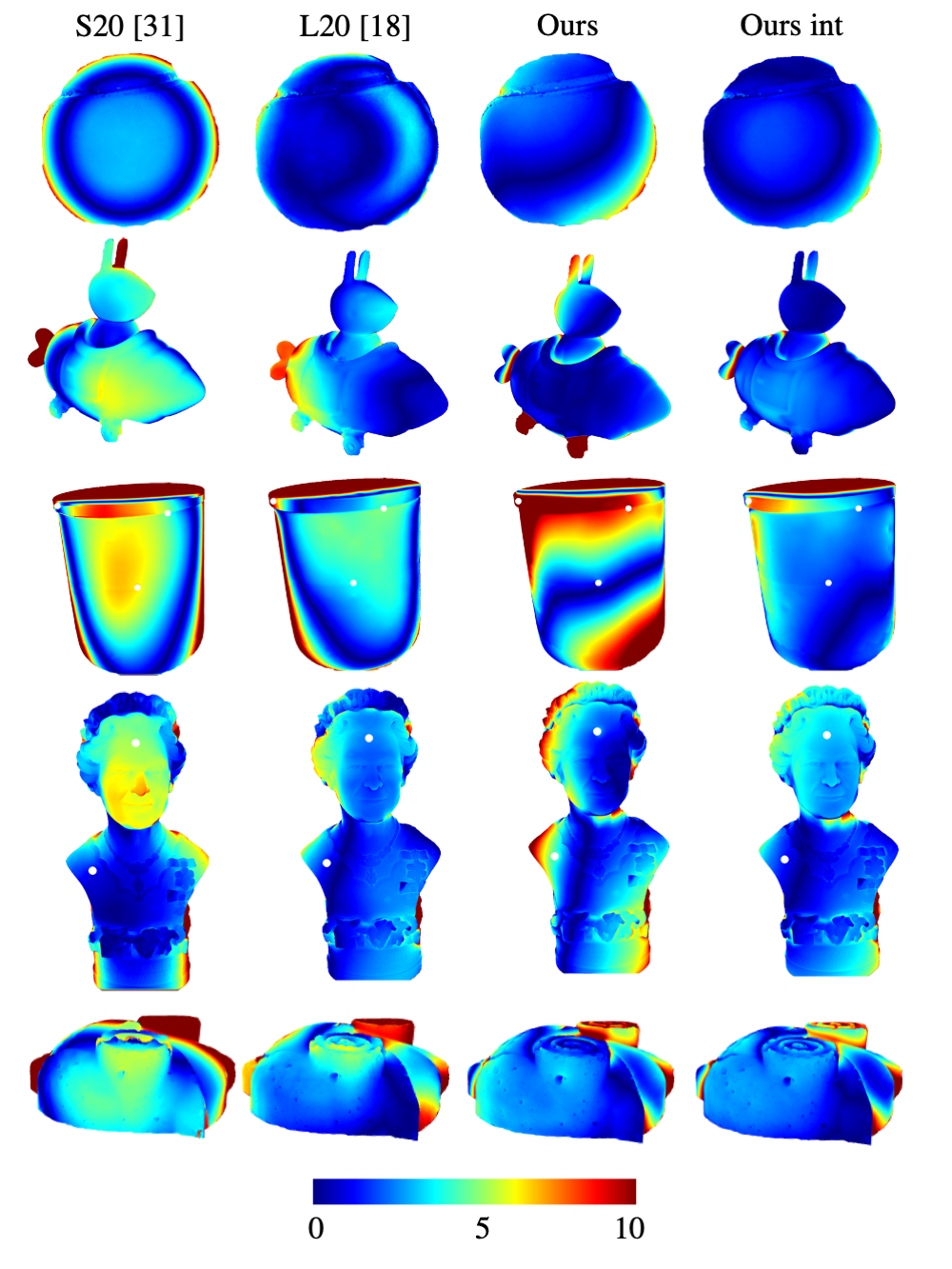}
    \vspace{-4mm}
    \caption{Depth error maps (in mm) on the LUCES dataset in the calibrated case}
    \label{fig:depth_err_cal}
\end{figure*}

\textbf{Additional Calibrated Results}
In Fig. \ref{fig:normal_err_cal} we present some additional normal maps from our tests on the LUCES dataset in the calibrated case. In Fig. \ref{fig:depth_err_cal} we show the depth error maps for each method in the calibrated case.

\begin{figure}[!h]
    \centering
    \includegraphics[width=\columnwidth]{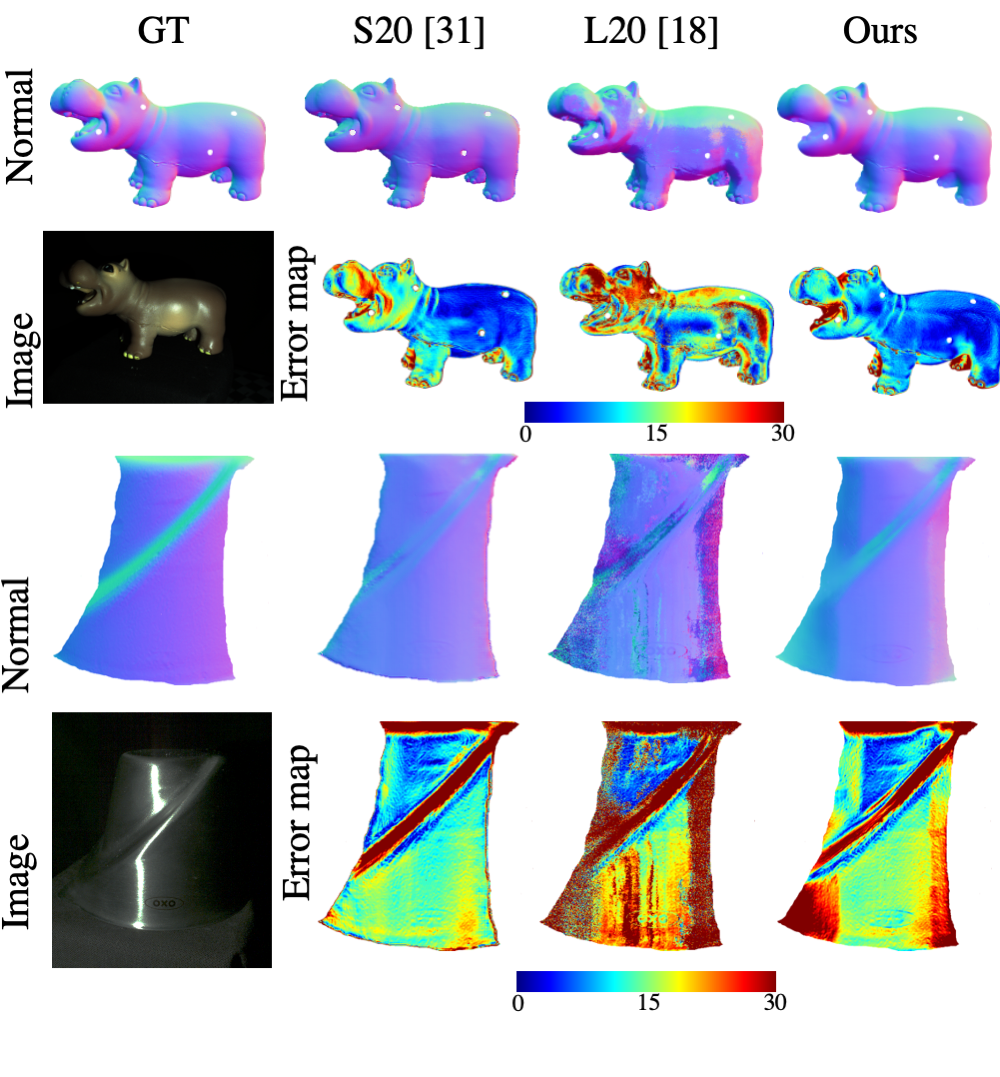}
    \caption{Additional normal predictions and error maps (in degrees) on the LUCES dataset in the uncalibrated case.}
    \label{fig:normal_err_uncal}
\end{figure}

\begin{figure*}[!h]
    \centering
    \includegraphics[width=0.9\textwidth]{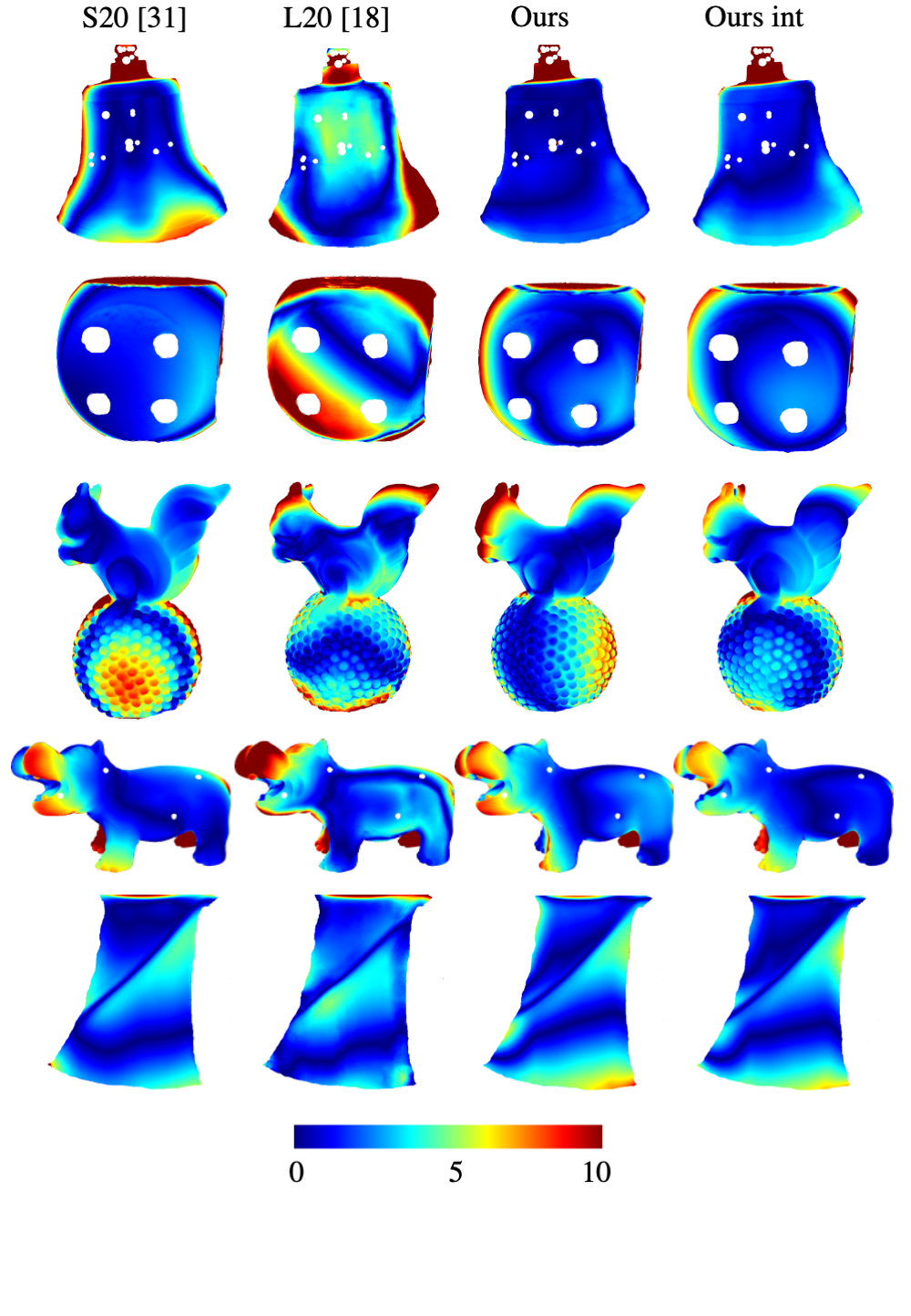}
    \vspace{-23mm}
    \caption{Depth error maps (in mm) on the LUCES dataset in the uncalibrated case}
    
    \label{fig:depth_err_uncal}
\end{figure*}

\textbf{Additional Uncalibrated Results}
In Fig. \ref{fig:normal_err_uncal} we present some additional normal maps from our tests on the LUCES dataset in the uncalibrated case. In Fig. \ref{fig:depth_err_uncal} we show the depth error maps for each method in the uncalibrated case.

\end{document}